\documentclass{article}

\usepackage[preprint]{neurips_2026}

\usepackage[utf8]{inputenc}  
\usepackage[T1]{fontenc}     
\usepackage{hyperref}        
\usepackage{url}             
\usepackage{booktabs}        
\usepackage{multirow}        
\usepackage{float}           
\usepackage{amsfonts}        
\usepackage{nicefrac}        
\usepackage{microtype}       
\usepackage{xcolor}          
\usepackage{graphicx}        
\usepackage{subcaption}      
\usepackage{amsmath}         
\usepackage[nameinlink]{cleveref}        

\title{Matryoshka Language Model Suites}

\author{%
  Nathan Godey \\
  Cornell University\\  
  \texttt{nathan.godey@cornell.edu} \\
  \And
  Yoav Artzi\\
  Cornell University\\
  \texttt{yoavartzi@cornell.edu}
}

\begin{document}

\maketitle

\begin{abstract}
Training a language model suite classically requires training each model separately and serving them independently. 
We improve both training and inference efficiency by stacking sub-models of increasing size into a single nested architecture trained end-to-end. 
This \emph{Matryoshka} training framework reduces the total parameter count of the suite, enables low-cost distillation from the largest to all smaller sub-models at every training step, and is well-suited for speculative decoding as the draft model is contained within the verifier. 
We validate our approach by training a Matryoshka suite comprising 500M, 1.5B, and 3B sub-models. Our suite is on par with independently trained baselines on benchmark performance and validation and out-of-domain perplexities, while using 36\% less training compute and improving the throughput of speculative decoding by 14--26\%.
We also ablate key architectural choices, offering guidance for building strong Matryoshka LM suites.
\end{abstract}

\section{Introduction}

\begin{figure}[!b]
  \centering
  \begin{subfigure}[h]{0.38\textwidth}
    \centering
    \includegraphics[width=\linewidth]{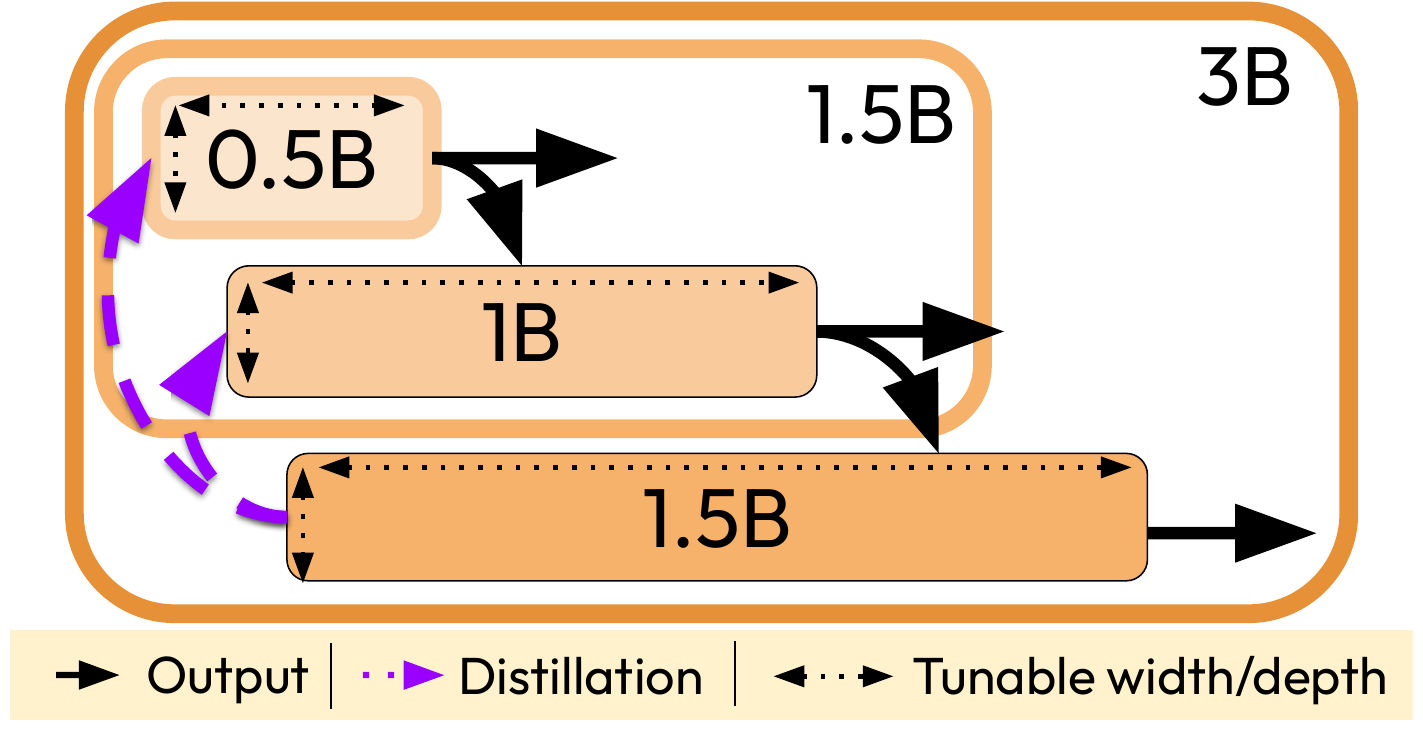}
    \caption{Matryoshka suite}
    \label{fig:stacked}
  \end{subfigure}
  \hfill
  \begin{subfigure}[h]{0.6\textwidth}
    \centering
    \raisebox{0pt}[\height][0pt]{\includegraphics[height=1em]{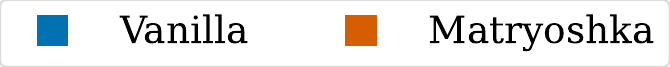}}\\[0.2em]
    \includegraphics[width=\linewidth]{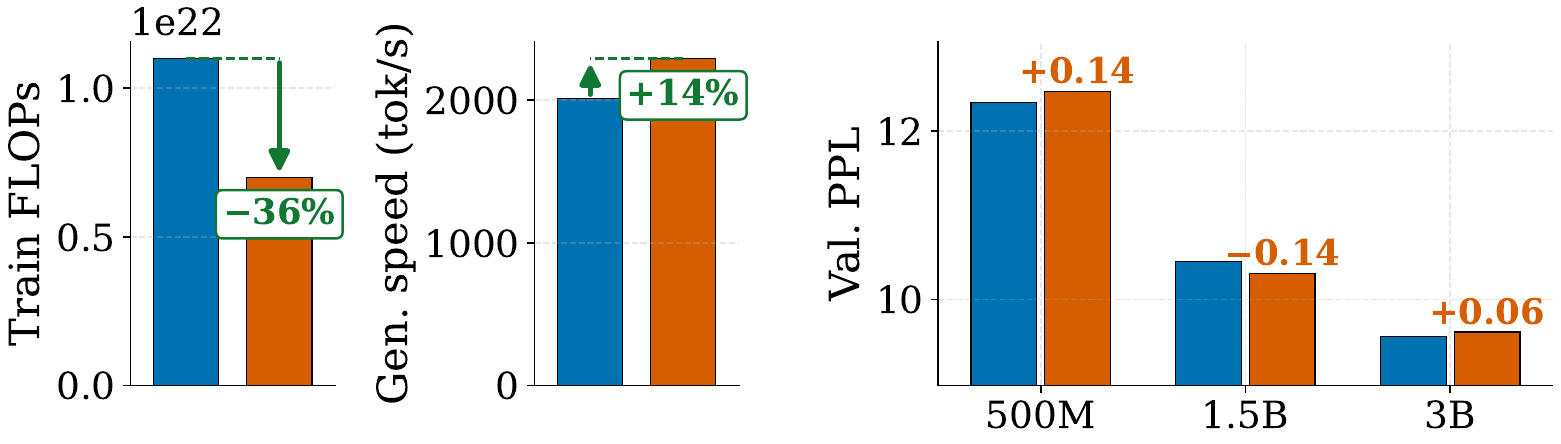}
    \caption{Main results for the 3B suite}
    \label{fig:headline-bars}
  \end{subfigure}
  \caption{\textbf{(a)} Matryoshka model suite: sub-models are nested into a single architecture
    trained end-to-end, instead of training $M$ models independently.
    Each sub-model has its own width and depth, fixed at training time;
    smaller sub-models are distilled from the largest one as a by-product
    of every forward pass.
    \textbf{(b)} Main results for the 3B suite: training compute drops by
    36\%, speculative-decoding throughput on a 500M-draft / 3B-verifier
    pair (nucleus, $T{=}0.4$) rises by 14\%, and per-size validation PPL
    stays within $\pm 1.4\%$ of the Vanilla baseline.}
  \label{fig:overview}
\end{figure}
\makeatletter\@notice\makeatother

The way training and inference costs of large language models (LLMs) scale steeply with
model size creates a demand  for models across a broad spectrum of
sizes, from 100M~\citep{allal2025smollm2smolgoesbig} to
trillion~\citep{deepseekai2026deepseekv4} parameters, for diverse
deployment scenarios. Delivering models from different scales allows for different load/performance tradeoffs, with smaller models being more memory- and compute-efficient than their larger counterparts. As a result, models are usually released in \emph{suites}: groups of
models trained separately on similar data, with smaller models often distilled
from larger ones in the suite. The question of efficiency in training and serving
model suites \emph{jointly} is rarely studied. We propose a training framework
that is particularly well-suited to the model-suite paradigm and offers several
concrete benefits over the standard approach.

Instead of training separate models, our approach progressively stacks the
outputs of smaller sub-models as inputs to larger ones. For instance,
rather than training a $\{1\text{B},\,8\text{B},\,30\text{B},\,70\text{B}\}$
suite as four independent runs, our method trains a single
architecture combining a $1$B sub-model with additional $7$B, $22$B, and
$40$B parameter blocks, resulting in 70B trained parameters instead of 109B. Contrarily to previous work \citep{devvrit2023matformer}, each sub-model can be detached as a standalone
language model, with its own depth and KV cache footprint, and served independently just like a regular checkpoint.
This setup also makes distillation from the largest model to all smaller ones a natural
by-product, as logits from all submodels are available during training. As a result, training the Matryoshka
suite costs on the same order as training only the largest model alone,
and delivers all $N$ sub-models in a single run, with cheap knowledge distillation.

At inference time, the nested structure further benefits speculative
decoding~\citep{specdec}: the draft model is nested in the verifier model, and both already share KV cache, eliminating the usual memory overhead of
maintaining a separate draft model. Cross-model prediction agreement is also
higher than in independently trained suites, directly increasing token
acceptance rates.

We verify our approach by training a 3B-parameter Matryoshka suite with 500M, 1.5B, and 3B sub-models on 35B tokens from the FineWeb-Edu dataset. Our suite matches independently trained baselines on validation perplexity, out-of-domain perplexity, and benchmark performance while using 36\% less total training compute. Moreover, the shared-weight structure and low-cost distillation substantially improves cross-model agreement and makes larger draft models practical for speculative decoding. We also conduct an extensive analysis of architectural design, and show the impact of width and depth configurations in Matryoshka suites, and that both the distillation loss and our model junction mechanism are crucial for the performance of smaller sub-models. We release trained checkpoints and a HuggingFace's \texttt{transformers}-compatible implementation\footnote{Checkpoints \& inference code: \url{https://huggingface.co/nthngdy/matryoshka-3B}}.

Overall, our contributions can be summarized as:
\begin{itemize}
  \item We introduce \emph{Matryoshka Language Model Suites}, a pretraining
    framework that nests sub-models of increasing width and depth into a single
    jointly trained architecture (\Cref{sec:method});
  \item We derive a low-cost distillation objective that leverages the
    availability of all sub-model logits at every forward pass;
  \item We train a 3B-parameter Matryoshka suite and show it reaches
    near-parity (within 0.5 average points at every size) with independently trained baselines on standard benchmarks
    while using 36\% less training compute (\Cref{sec:experiments});
  \item We demonstrate that the shared-weight structure and low-cost distillation substantially improve cross-model agreement and make larger drafts practical for speculative decoding, where plain vanilla speculative decoding barely improves
    over autoregressive inference.
  \item We provide extensive ablations at smaller scale (50M/100M/200M suite) that provide insights for architectural hyperparameters and training objectives.
\end{itemize}
\section{Method}
\label{sec:method}
We nest smaller models into larger ones, and allow different hidden sizes across models by implementing a simple junction mechanism. We then compute a cross-entropy loss and a largest-model distillation loss for each exit points, and use a convex combination of these as a single overall objective.

\paragraph{Modeling}
A Transformer~\citep{NIPS2017_3f5ee243} language model maps an input sequence
$x \in [1,V]^L$ to output representations $o_\theta(x) \in \mathbb{R}^{L \times D}$
via an input embedding $e_\theta$ and a stack of Transformer blocks $T_\theta$ of
hidden dimension $D$. A linear LM head $W$ maps each position's representation
$o_\theta(x)$ to logits $l \in \mathbb{R}^V$, and the parameters $(\theta, W)$
are trained to minimize the next-token prediction cross-entropy loss.\footnote{We notationally distinguish the LM head parameters to simplify the definition of our Matryoshka suite.}

A conventional suite of $M$ models uses $M$ independent parameter sets
$\{(\theta_m, W_m)\}_{m=1}^M$. We instead enforce a strict nesting of parameters with the exception of the LM head:
\[
  \theta_1 \subset \theta_2 \subset \cdots \subset \theta_M = \theta.
\]
Each sub-model $T_{\theta_m}$ is a separate Transformer stack with its own hidden dimension (or width) $D_m$, number of layers (or depth) $n_m$, and a LM head $W_m$.
We require strictly increasing
hidden dimensions $D_1 \leq D_2 \leq \cdots \leq D_M$. Unlike standard early
exiting, which assumes a fixed hidden dimension across exits, the
Matryoshka design lets each sub-model use a width tailored to its size,
giving more flexibility in the architectural design at every sub-model size compared to early-exit methods~\citep{elhoushi2024layerskip}.
Moreover, because $\theta_m \subset \theta_{m+1}$, running the full architecture
produces all $M$ sub-model logits in a single forward pass while
activating fewer parameters and writing less KV cache than $M$ independent
forward passes, facilitating distillation during training and multi-model inference methods such as speculative decoding.

\paragraph{Inter-model Junction}
The main modeling difficulty consists in passing the output $o^{m}_\theta(x) \in
\mathbb{R}^{D_m}$ of sub-model $T_{\theta_m}$ as the input to sub-model $T_{\theta_{m{+}1}}$, which
operates in $\mathbb{R}^{D_{m+1}}$ with $D_{m+1} \geq D_m$. We design a junction mechanism that does not introduce any additional parameters.

The most naive approach is to
concatenate $o^{m}_\theta(x)$ with a fresh embedding $e^{m+1}_\theta(x) \in
\mathbb{R}^{D_{m+1}-D_m}$ extending to the ``new'' dimensions that sub-model
$m{+}1$ adds. However, this creates a magnitude mismatch: Transformer outputs
typically have much larger norms than input embeddings, which destabilizes
training in the low-index channels. We address this by rescaling
$o^{m}_\theta(x)$ so that its norm matches the norm of $e^{m+1}_\theta(x)$
before the concatenation.
Formally, given $o^{m}_\theta(x)$ the output embedding coming from the $m$-th submodel and $e^{m+1}_\theta(x)$ the input embedding of the $(m+1)$-th submodel, we build the $(m+1)$-th submodel output $o^{m+1}_\theta(x)$ as follows:
\begin{equation}
  \begin{aligned}
    &\tilde{o}^{m}_\theta(x) = o^{m}_\theta(x)\cdot
      \dfrac{\|e^{m+1}_\theta(x)\|_2}{\|o^{m}_\theta(x)\|_2} \\[8pt]
    &o^{m+1}_\theta(x) =
      T^{m+1}_\theta\!\left(\operatorname{concat}\!\left(
        e^{m+1}_\theta(x),\;\tilde{o}^{m}_\theta(x)
      \right)\right)\;\;.
  \end{aligned}
  \label{eq:junction}
\end{equation}
The first
$D_m$ channels of the input to $T^{m+1}_\theta$ carry information from sub-model
$m$, while the remaining $D_{m+1} - D_m$ channels are freshly initialized,
analogous to the input embedding of a standalone model. Each
sub-model has a dedicated LM head $W^m_\theta \in \mathbb{R}^{V \times D_m}$
and is trained with its own cross-entropy loss $\mathcal{L}^m_{\mathrm{ce}}$.
Both design choices matter empirically: dropping the norm rescaling, or
zeroing out the fresh embedding, degrades downstream PPL across all
sub-model sizes (\Cref{tab:junction}).

\paragraph{Integrated Distillation}
In classical suite training, distillation from the largest model into
smaller ones~\citep{peng2025pretrainingdistillation,gu2025miniplm} requires
either storing teacher logits offline or running the teacher in parallel
during student training. Each option adds significant compute or storage
cost. In a Matryoshka suite, every forward pass through $\theta_M$ already
produces logits for all $M$ sub-models simultaneously, making online
distillation from the largest sub-model to all smaller ones free.

Concretely, we add a distillation term to each sub-model $m < M$, defined as
the cross-entropy between the teacher and student next-token distributions:
\begin{equation}
  \mathcal{L}^{M \to m}_d =
    - \sum_{v=1}^{V} \texttt{stop\_grad}(\sigma(l^M_t)_v) \, \log \sigma(l^m_t)_v,
  \label{eq:distillation}
\end{equation}
where $\sigma$ denotes the softmax. The total loss for sub-model $m$ is
a convex combination:
\begin{equation}
  \mathcal{L}^m = (1 - \alpha_d)\,\mathcal{L}^m_{\mathrm{ce}}
                 + \alpha_d\,\mathcal{L}^{M \to m}_d,
  \label{eq:submodel-loss}
\end{equation}
and the overall training objective sums over sub-models: $\mathcal{L} = \sum_{m=1}^M \mathcal{L}^m$.
The distillation coefficient $\alpha_d \in [0,1]$ controls the strength of the
teacher signal. We study its effect in \Cref{sec:appendix-distil}. We observe that $\alpha_d$ should be set to lower values compared to offline setups.

\paragraph{Speculative Decoding}
At inference time, any pair $(m, m')$ with $m < m'$ forms a natural
draft-verifier pair for speculative decoding~\citep{specdec}. Sub-model $m$
generates a draft sequence of $\gamma$ tokens, which sub-model $m'$ then
verifies in a single forward pass. The Matryoshka structure is orthogonal
to draft-acceleration techniques such as EAGLE~\citep{li2024eagle} or
MEDUSA~\citep{cai2024medusa}, which can be applied on top to further speed
up inference. What we explore here is the complementary benefit of weight
and KV-cache sharing between draft and verifier, which makes larger and
hence stronger draft models practical.

Two properties make Matryoshka suites particularly well-suited for this setup.
Because $\theta_m \subset \theta_{m'}$, the verifier's first
$n_1 + \cdots + n_m$ layers coincide with the draft model's layers. The KV cache
computed during drafting is fully reused by the verifier for those shared
layers; only the $n_{m+1} + \cdots + n_{m'}$ additional layers need to run.
This eliminates the memory cost of maintaining a separate draft model.
Draft-verifier alignment is higher than in independently trained suites
of the same sizes, as we confirm in \Cref{sec:experiments}, which directly
translates to a higher token acceptance rate.

Together, these two properties make non-tiny drafts practical. Standard
speculative decoding favors draft models one or two orders of magnitude
smaller than the verifier (e.g., 60M drafts for an 11B
target~\citep{specdec}, or 160M--1B drafts for 7B--70B
Llama~\citep{zhou2024distillspec,li2024eagle}); larger drafts typically
do not amortize, since their drafting cost and KV cache footprint grow
faster than their acceptance rate. We show in \Cref{sec:exp:spec} that a 500M / 3B pair (1:6 size ratio) is
already in the unfavorable regime, with speculative decoding degrading latency compared to plain autoregressive
inference. Yet, the Matryoshka 500M / 3B pair turns the same configuration into a 20--40\% speedup over its own standard decoding by sharing its KV cache and early layers.

\section{Experiments}\label{sec:experiments}

\subsection{Experimental Setup}\label{sec:exp:setup}

\paragraph{Architecture}
We train a Matryoshka suite from scratch with three nested sub-models
totalling 3B parameters, and compare it to an equivalent traditional suite;
configurations are summarized in \Cref{tab:arch}. The main architectural
design choice is how to allocate the width and depth to each sub-model while
matching the target parameter count.\footnote{Given a target parameter count, and using standard configurations for FFN layers ($D_m \rightarrow 4\cdot D_m \rightarrow D_m$), fixing the depth $n_m$ of each sub-model determines its width $D_m$.} We adopt a deployment-oriented viewpoint
and aim to match both memory and compute footprints across suite types.

Because Matryoshka sub-models combine blocks of different widths, their
compute and memory trade-offs differ from a uniform-width architecture, and
the depth budget that balances per-token FLOPs and KV cache shifts
accordingly. We use simple closed-form formulas for parameter count, KV
cache, and per-token inference FLOPs (\Cref{sec:appendix-accounting}). In our experiments, we use $L_v = 28$ layers for the standard 3B model and find $L = 39$ total layers to be the sweet spot for the
3B Matryoshka suite: shallower budgets (e.g.\ $L = 28$ matching the Vanilla baseline)
reduce memory per token below the Vanilla footprint, while deeper ones
(e.g.\ $L = 50$) inflate KV cache, with both mismatching the footprint of
the Vanilla suite. 
\Cref{sec:appendix-layer-budget} provides further discussion and analysis.

\begin{figure}[tbp]
    \centering
    \hfill
    \begin{subfigure}[b]{0.45\textwidth}
        \includegraphics[width=\textwidth]{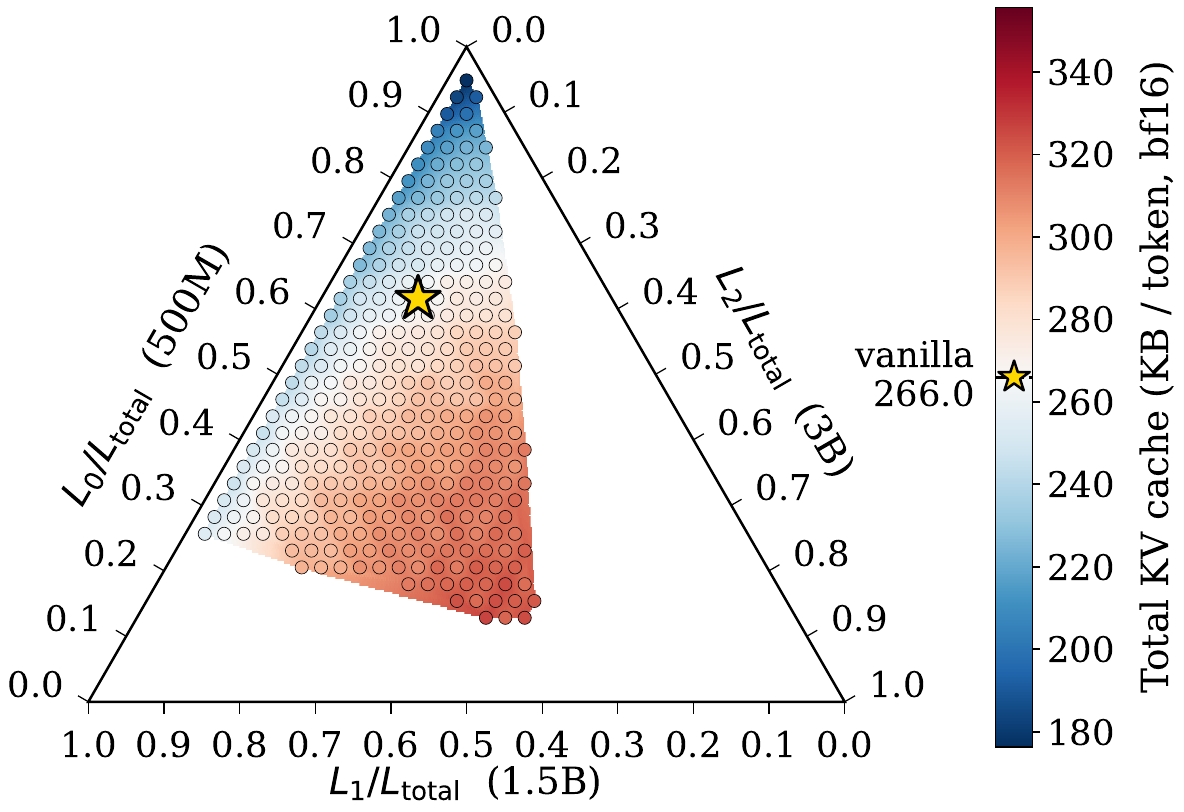}
        \caption{KV cache per token}
        \label{fig:sweep_3B_kvc}
    \end{subfigure}
    \hfill
    \begin{subfigure}[b]{0.45\textwidth}
        \includegraphics[width=\textwidth]{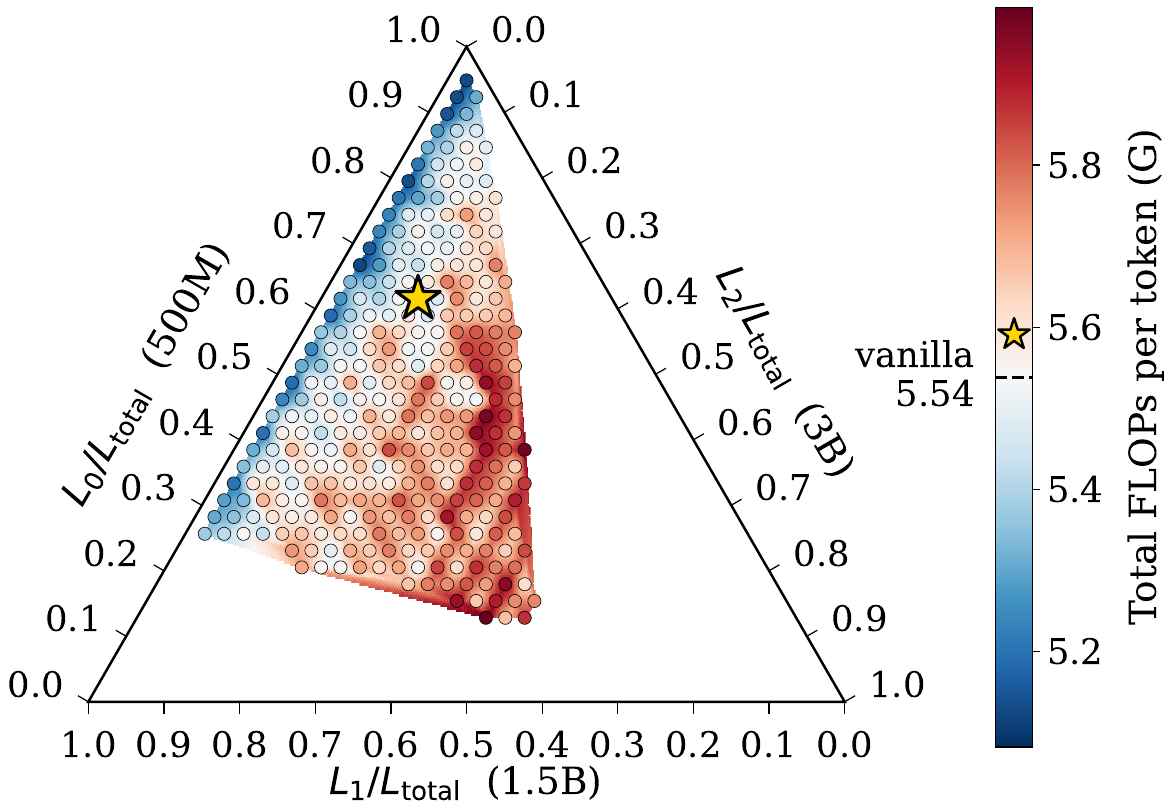}
        \caption{Theoretical FLOPs per token}
        \label{fig:sweep_3B_flops}
    \end{subfigure}
    \hfill\null
    \caption{Ternary sweep over all feasible depth triplets $(n_1, n_2, n_3)$
      summing to 39 layers for a (500M, 1.5B, 3B) Matryoshka suite (see
      \Cref{sec:appendix-ternary} for a guide to reading these plots).
      Each point's position encodes the fraction of total layers assigned to
      each sub-model; its color encodes KV cache (left) or FLOPs (right)
      for the 3B forward pass relative to the vanilla 3B baseline
      (dashed line, 266.0\,KB/token and 5.54\,GFLOPs/token).
      Blue points lie below the vanilla baseline; red points lie above.
      The gold star marks our chosen configuration $(24, 10, 5)$,
      which sits near the footprint-equivalent frontier.}
    \label{fig:sweep_3B}
\end{figure}

Within the $L = 39$ budget, we sweep all feasible depth triplets
$(n_1, n_2, n_3)$ summing to 39, fixing target sizes (500M, 1.5B, 3B) and
attention head dimensions (64, 96, 128), and for each triplet solve for the head
count that best matches each target. \Cref{fig:sweep_3B} reports the
resulting KV cache and per-token FLOPs of the full 3B forward pass relative
to a Vanilla 3B baseline. Our chosen configuration
$(n_1, n_2, n_3) = (24, 10, 5)$ closely matches both KV cache and per-token FLOPs to the Vanilla 3B baseline. We notably choose to use the same width and depth for the 500M sub-model in both suites, to provide a strictly comparable data point.
All sub-models share the SmolLM2 tokenizer~\citep{allal2025smollm2smolgoesbig}
(49{,}152 tokens) and use RoPE~\citep{rope} positional embeddings
with $\theta_{\mathrm{RoPE}} = 10^5$.

\begin{table}[t]
\centering\small
\begin{tabular}{llrrrrrrr}
\toprule
 & & Params & Cumul. & $D_m$ & Layers & Heads & Head dim & Int. size \\
\midrule
\multirow{3}{*}{Vanilla}
  & 500M  & \multicolumn{2}{c}{0.50B} & 1\,024 & 24 & 16 & 64  & 4\,096  \\
  & 1.5B  & \multicolumn{2}{c}{1.51B} & 1\,728 & 28 & 18 & 96  & 6\,912  \\
  & 3B    & \multicolumn{2}{c}{3.19B} & 2\,560 & 28 & 20 & 128 & 10\,240 \\
\midrule
\multirow{3}{*}{Matryoshka}
  & 500M  & 0.50B & 0.50B & 1\,024 & 24 & 16 & 64  & 4\,096  \\
  & 1.5B  & 0.98B & 1.48B & 2\,304 & 10 & 24 & 96  & 9\,216  \\
  & 3B    & 1.72B & 3.20B & 4\,352 &  5 & 34 & 128 & 17\,408 \\
\bottomrule
\end{tabular}
\caption{Architecture of the Vanilla baseline and the 3B Matryoshka suite.
\emph{Params} is the incremental block size; \emph{Cumul.} is the running
total reached at each exit point. The suite totals are 5.2B parameters for
Vanilla and 3.2B for Matryoshka ($-38\%$), at matched exit sizes.}
\label{tab:arch}
\end{table}

\paragraph{Training Setup}
Training uses 35B tokens from
FineWeb-Edu~\citep{penedo2024the}, a high-quality education-focused web
corpus, with sequences packed to length 2{,}048. The optimizer is AdamW with
$(\beta_1, \beta_2) = (0.9, 0.95)$, $\epsilon = 10^{-8}$, peak learning rate
$4 \times 10^{-4}$, weight decay $0.01$, and gradient clipping at $1.0$.
We use a warmup-stable-decay (WSD) schedule with 3{,}000 cooldown steps over
33{,}000 total steps, a batch size of 512 sequences, and bf16 mixed precision.
The distillation coefficient is set to $\alpha_d = 0.3$ for all main-suite
runs, chosen from a smaller-scale proxy sweep
(\Cref{sec:appendix-distil}).
Combined training across both suites amounts to 52 GPU-days on NVIDIA B200.

As our primary baseline we train a Vanilla suite of three independent
Llama-style~\citep{llama3modelcard} models at roughly the same sizes
(0.50B, 1.51B, 3.19B; see \Cref{tab:arch}), using the same tokenizer,
data, and hyperparameters as the Matryoshka suite. Their shapes follow
established small-LM conventions: the 500M and 1.5B Vanilla baselines
mirror the SmolLM2~\citep{allal2025smollm2smolgoesbig} configurations at
those sizes, and the 3B Vanilla baseline follows the Llama-3.2-3B shape.
Both suites are trained for 35B tokens; we additionally cool down and
evaluate Vanilla checkpoints at 23B tokens to obtain a compute-matched
reference.

\subsection{Results and Analysis}\label{sec:exp:results}

We evaluate the Matryoshka suite against the Vanilla baseline along three
axes: benchmark accuracy (\Cref{sec:exp:bench}), training dynamics and
cross-model alignment (\Cref{sec:exp:dynamics}), and speculative decoding
throughput (\Cref{sec:exp:spec}). \Cref{sec:exp:ablations} reports extensive
ablations for widths and depths configurations, inter-model junction and loss coefficients at a smaller scale (50M/100M/200M parameters suite trained on 20B tokens).

\subsubsection{Downstream Evaluation}\label{sec:exp:bench}

We evaluate both suites zero-shot on seven multiple-choice benchmarks
that are standard at this scale: ARC-Easy and
ARC-Challenge~\citep{allenai:arc}, HellaSwag~\citep{zellers2019hellaswag},
LAMBADA~\citep{paperno2016lambada}, OpenBookQA~\citep{obqa},
PIQA~\citep{piqa}, and Winogrande~\citep{winogrande}.
We also report average byte perplexity on five out-of-distribution corpora
(WikiText-103~\citep{w103}, C4~\citep{c4},
PG-19~\citep{pg19}, arXiv, and PubMed); the per-corpus values are reported in
\Cref{sec:appendix-ood-ppl}.

\begin{figure}[tbp]
    \centering
    \includegraphics[height=1.2em]{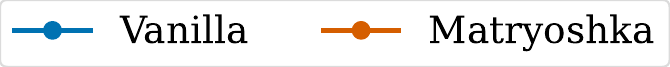}

    \begin{subfigure}[b]{0.32\textwidth}
        \includegraphics[width=\textwidth]{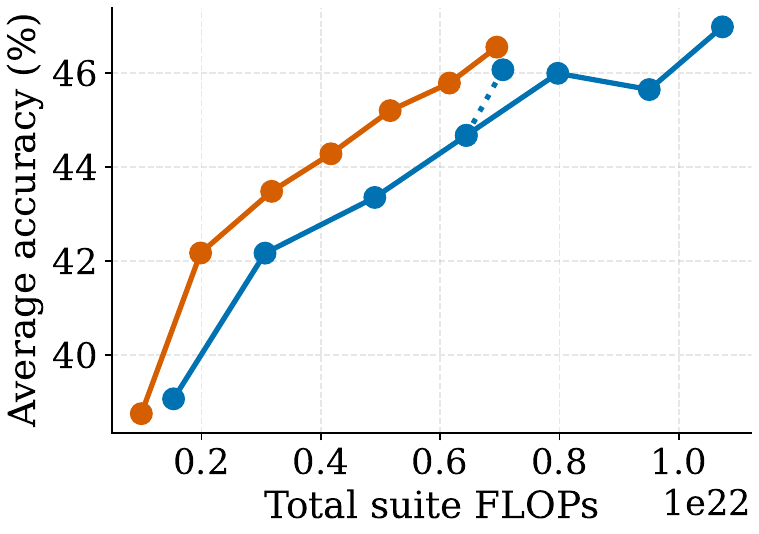}
        \caption{500M}
        \label{fig:bench_500m}
    \end{subfigure}
    \hfill
    \begin{subfigure}[b]{0.32\textwidth}
        \includegraphics[width=\textwidth]{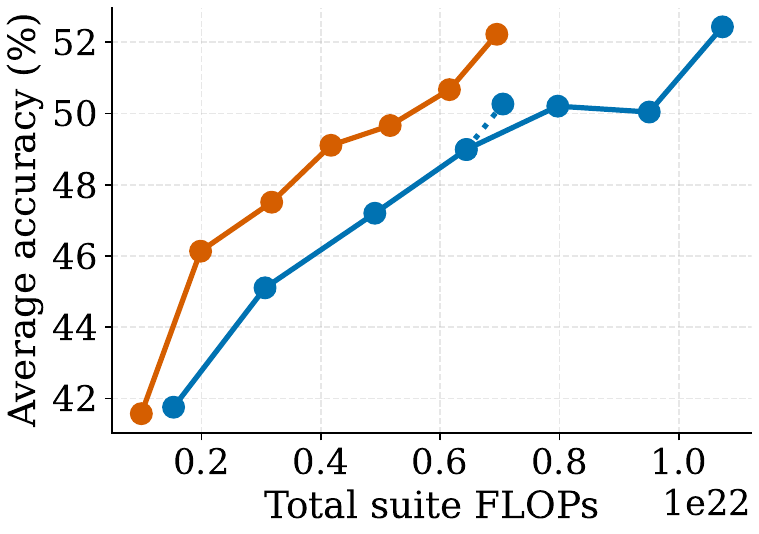}
        \caption{1.5B}
        \label{fig:bench_1-5b}
    \end{subfigure}
    \hfill
    \begin{subfigure}[b]{0.32\textwidth}
        \includegraphics[width=\textwidth]{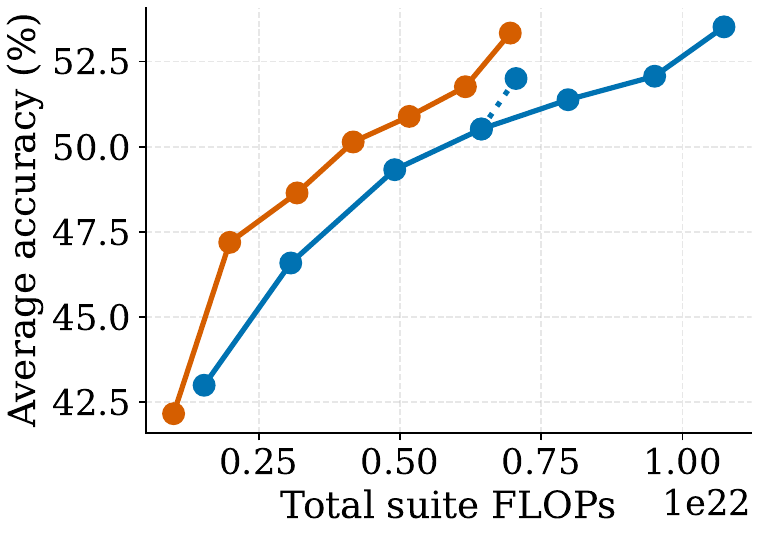}
        \caption{3B}
        \label{fig:bench_3b}
    \end{subfigure}
    \caption{Average benchmark accuracy as a function of total suite training
      FLOPs for each sub-model size. The Matryoshka curve lies above the
      compute-matched Vanilla baseline at every sub-model size, with the
      largest gap at 1.5B. The dotted segment at the end of the vanilla
      curve is a 2{,}000-step learning-rate cooldown branched off the
      21B-token checkpoint, reaching 23B tokens.}
    \label{fig:triple}
\end{figure}

\Cref{tab:benchmarks} shows that the Matryoshka suite reaches near-parity with the
Token-matched Vanilla baseline on average accuracy at every size (within
0.5 points) while using 36\% less training compute, and outperforms the
FLOPs-matched Vanilla baseline at every size by +0.4 to +1.9 points. On
average, OOD byte perplexity, the Matryoshka suite outperforms the
Token-matched Vanilla baseline at the 1.5B and 3B scales (and ties at
500M), with the largest single-corpus advantage on arXiv
(\Cref{sec:appendix-ood-ppl}).
\Cref{fig:triple} shows the same comparison as a function of total suite
training FLOPs. The Matryoshka curve dominates the Vanilla curve at every
compute level for every sub-model size.

\begin{table}[t]
\centering \small
\resizebox{\textwidth}{!}{%
\begin{tabular}{lccccccccc}
\toprule
Model & \multicolumn{1}{c}{ARC-E} & \multicolumn{1}{c}{ARC-C} & \multicolumn{1}{c}{HS} & \multicolumn{1}{c}{Lambada} & \multicolumn{1}{c}{OBQA} & \multicolumn{1}{c}{PIQA} & \multicolumn{1}{c}{WG} & \multicolumn{1}{c}{\textbf{Avg acc.} $\uparrow$} & \multicolumn{1}{c}{\textbf{OOD PPL} $\downarrow$} \\
\midrule
\multicolumn{10}{l}{\textit{500M parameters}} \\
\midrule
Vanilla (=FLOPs) & 56.9          & 31.2          & 43.2          & 36.5          & 34.4          & 68.8          & 51.5          & 46.1          & 2.295 \\
Vanilla (=Tokens) & \textbf{59.0} & \textbf{31.5} & \textbf{44.8} & \textbf{37.8} & \textbf{35.2} & 68.6          & 52.0          & \textbf{47.0} & \textbf{2.263} \\
Matryoshka              & 58.8          & 30.7          & 44.4          & 36.1          & 33.8          & \textbf{69.0} & \textbf{53.0} & 46.5          & 2.265 \\
\midrule
\multicolumn{10}{l}{\textit{1.5B parameters}} \\
\midrule
Vanilla (=FLOPs) & 62.2          & 34.6          & 50.3          & 42.6          & 37.6          & 71.8          & 52.7          & 50.3          & 2.171 \\
Vanilla (=Tokens) & \textbf{65.8} & 35.7          & \textbf{53.5} & 43.9          & 39.4          & \textbf{72.1} & \textbf{56.7} & \textbf{52.4} & 2.139 \\
Matryoshka              & 63.3          & \textbf{35.8} & 53.3          & \textbf{44.7} & \textbf{40.2} & 71.9          & 56.4          & 52.2          & \textbf{2.121} \\
\midrule
\multicolumn{10}{l}{\textit{3B parameters}} \\
\midrule
Vanilla (=FLOPs) & 63.2          & 35.8          & 53.7          & 44.4          & 37.8          & 72.4          & 56.8          & 52.0          & 2.143 \\
Vanilla (=Tokens) & 64.4          & 37.3          & \textbf{56.9} & \textbf{46.3} & 38.2          & \textbf{73.8} & \textbf{57.8} & \textbf{53.5} & 2.097 \\
Matryoshka              & \textbf{64.6} & \textbf{38.0} & 55.9          & 44.9          & \textbf{39.0} & 73.2          & 57.7          & 53.3          & \textbf{2.067} \\
\bottomrule
\end{tabular}
}
\caption{Per-benchmark accuracy and out-of-distribution perplexity.
Vanilla~(=FLOPs) uses the same $7.0\mathrm{e}21$ FLOPs as Matryoshka
(23B tokens); Vanilla~(=Tokens) uses the same 35B tokens at 57\% more
compute. Avg acc.\ averages the seven benchmarks (WG = Winogrande); OOD PPL
averages WikiText-103, C4, PG-19, arXiv, and PubMed Central
(\Cref{sec:appendix-ood-ppl}). Matryoshka reaches near-parity with
Token-matched on accuracy and beats it on OOD PPL at 1.5B and 3B.}\label{tab:benchmarks}
\end{table}

\subsubsection{Training Dynamics and Model Agreement}\label{sec:exp:dynamics}

\begin{figure}[t]
    \centering
    \includegraphics[height=1em]{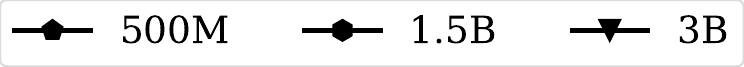}\hfill
    \includegraphics[height=1em]{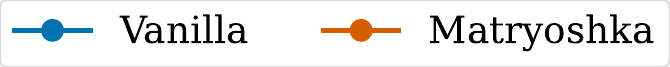}\hfill
    \includegraphics[height=1em]{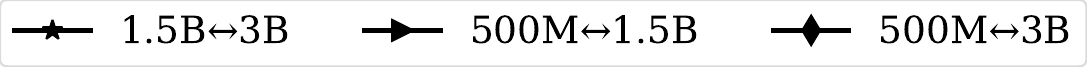}

    \begin{subcaptionblock}{0.32\linewidth}
        \includegraphics[width=\linewidth]{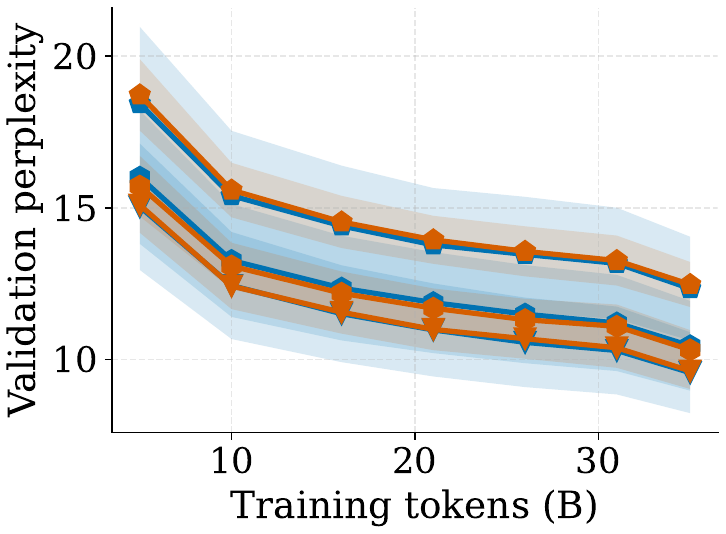}
        \caption{Perplexity}\label{fig:dynamics:ppl}
    \end{subcaptionblock}
    \hfill
    \begin{subcaptionblock}{0.32\linewidth}
        \includegraphics[width=\linewidth]{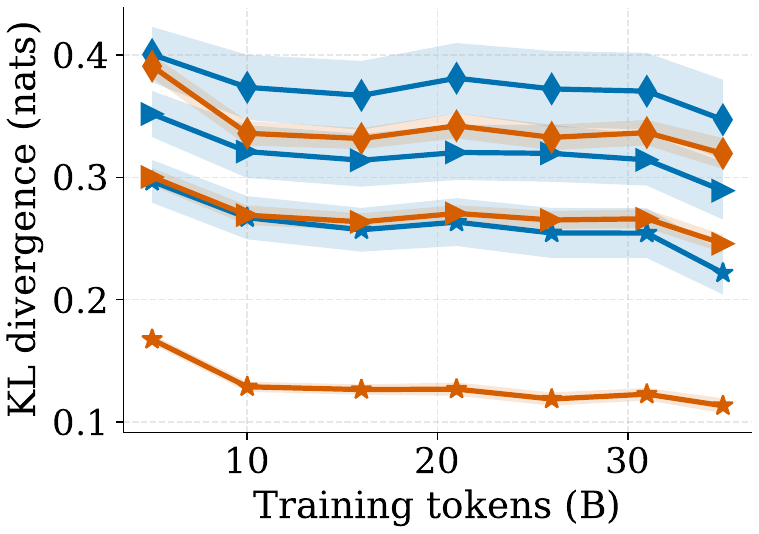}
        \caption{KL divergence}\label{fig:dynamics:kldiv}
    \end{subcaptionblock}
    \hfill
    \begin{subcaptionblock}{0.32\linewidth}
        \includegraphics[width=\linewidth]{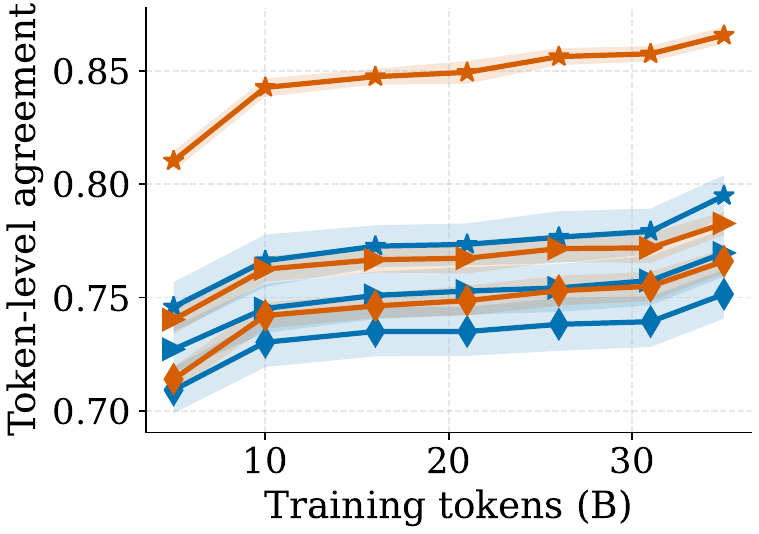}
        \caption{Agreement rate}\label{fig:dynamics:agreerate}
    \end{subcaptionblock}

    \caption{Training dynamics for the 3B vanilla and Matryoshka suites.
      Matryoshka matches vanilla perplexity throughout training while
      achieving significantly lower cross-model KL divergence and higher
      next-token agreement.}
    \label{fig:dynamics}
\end{figure}

\Cref{fig:dynamics} tracks validation perplexity, pairwise KL divergence,
and pairwise top-1 next-token agreement throughout training. Matryoshka
perplexity is on par with Vanilla perplexity at every checkpoint and every
size (\Cref{fig:dynamics:ppl}); the parameter sharing and the junction
mechanism do not harm language modeling quality. The small per-size gaps
that remain are not intrinsic to the nested architecture: a proxy-scale
sweep at 200M shows that simply rebalancing the per-sub-model loss
weights, currently uniform in our recipe, closes a substantial fraction of
the residual gap and shifts which size is the bottleneck
(\Cref{sec:appendix-loss-weighting}). 

A clear difference
appears in cross-model alignment, as Matryoshka pairs exhibit noticeably
lower KL divergence (\Cref{fig:dynamics:kldiv}) and higher
token-level agreement (\Cref{fig:dynamics:agreerate}) than independently
trained Vanilla pairs of matched sizes, with the effect strongest for the
(1.5B, 3B) pair (+5.7\% final agreement). This is a direct consequence of both the shared-weight
structure and of online distillation across training. Independently trained models have no such
constraint and develop different representations.
The next-token agreement rate is the quantity that governs speculative
decoding efficiency, which we discuss in \Cref{sec:exp:spec}.

\subsubsection{Speculative Decoding Throughput}\label{sec:exp:spec}

We measure speculative decoding throughput on an NVIDIA A100 GPU using the
500M sub-model as draft and the 3B sub-model as verifier, under both greedy
and nucleus ($p=0.9$) sampling. The 500M / 3B size ratio (1:6) is large by
literature standards, where draft models are typically one to two orders
of magnitude smaller than the
verifier~\citep{specdec,zhou2024distillspec,li2024eagle}. Each run uses
prompts streamed from FineWeb-Edu truncated to 1{,}024 tokens, and generates
512 new tokens per prompt, sampled with temperature $0.4$. The maximum batch
size that can fit the 80GB VRAM in such a setup is 64 for the Vanilla pair and 102 for the Matryoshka
pair. The Matryoshka setup fits a larger batch in the same KV-cache budget thanks to the
shared draft/verifier cache. A larger batch size allows for faster drafts relative to verification steps, as it reduces the bandwidth latency. Each metric is averaged over roughly 1M generated tokens.

\begin{figure}[t]
    \centering
    \includegraphics[height=1.1em]{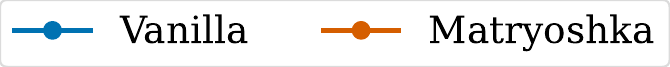}
    \includegraphics[height=1.1em]{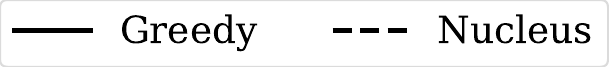}

    \hfill
    \begin{subcaptionblock}{0.4\linewidth}
        \includegraphics[width=\linewidth]{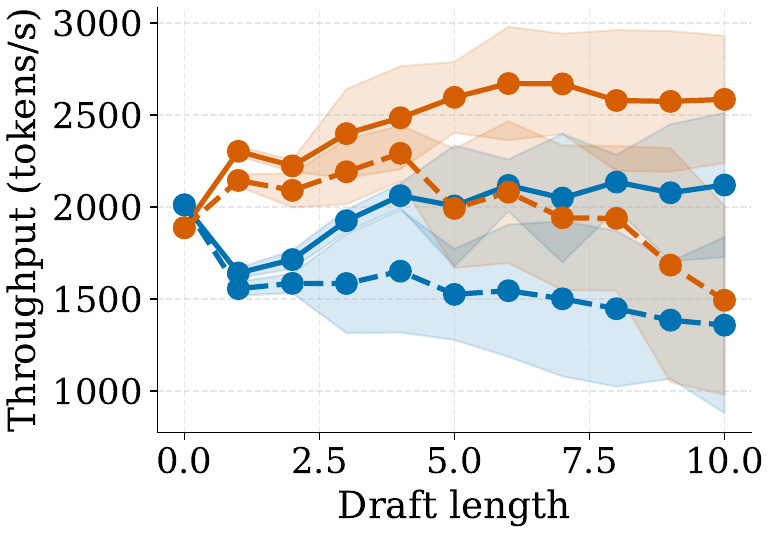}
        \caption{Throughput}
    \end{subcaptionblock}
    \hfill
    \begin{subcaptionblock}{0.4\linewidth}
        \includegraphics[width=\linewidth]{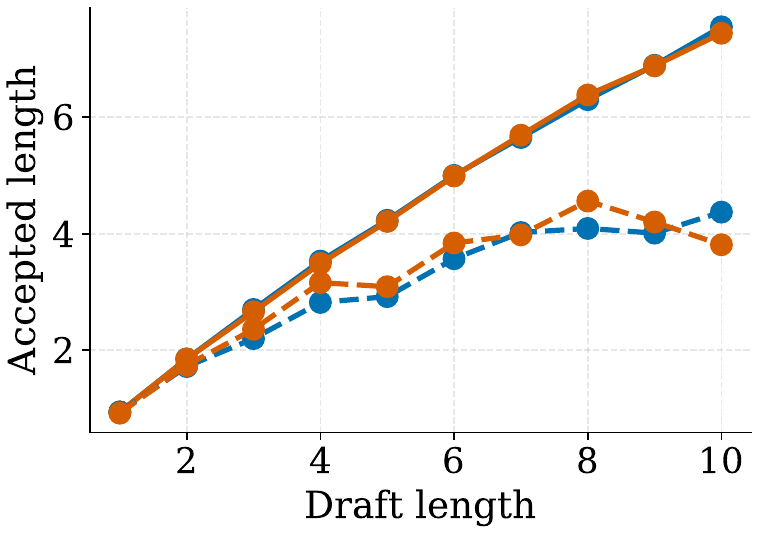}
        \caption{Accepted length}
    \end{subcaptionblock}
    \hfill\null

    \caption{Speculative decoding analysis with a 500M draft and 3B verifier.
      Solid lines show greedy decoding; dashed lines show nucleus sampling
      ($p=0.9$). The Matryoshka pair achieves higher throughput than the vanilla pair across draft lengths. In the nucleus case, the accepted length is generally higher for Matryoshka. Draft length of 0 corresponds to standard decoding.}
    \label{fig:spec}
\end{figure}

\Cref{fig:spec} shows that the Matryoshka pair achieves higher
accepted length than the Vanilla pair across most draft lengths in the nucleus case (+5\% in average), which
directly reflects the agreement-rate gap in
\Cref{fig:dynamics:agreerate}. The higher acceptance translates to higher
throughput: at a draft length of 6, Matryoshka reaches 2{,}650 tokens/s in
greedy mode against 2{,}100 for Vanilla (a 26\% speedup), and the gain is
preserved under nucleus sampling. Crucially, the Vanilla 500M--3B pair
barely beats plain autoregressive inference and is actually slower under
nucleus sampling. At this size ratio, the draft-side KV cache and drafting
FLOPs do not amortize against the acceptance rate.

\subsubsection{Comparison with MatFormer}

\begin{figure}[tbp]
    \centering
    \begin{subfigure}[b]{0.42\textwidth}
        \includegraphics[width=\textwidth]{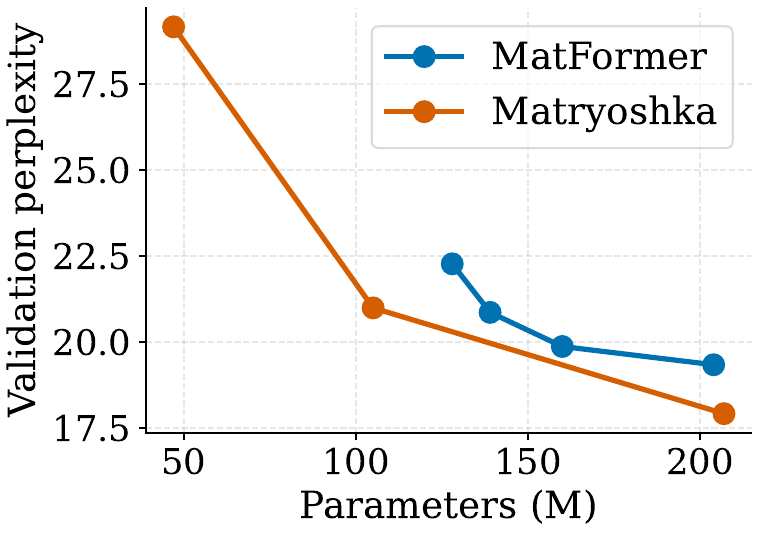}
        \caption{Validation perplexity}
        \label{fig:matformer:ppl}
    \end{subfigure}
    \hfill
    \begin{subfigure}[b]{0.42\textwidth}
        \includegraphics[width=\textwidth]{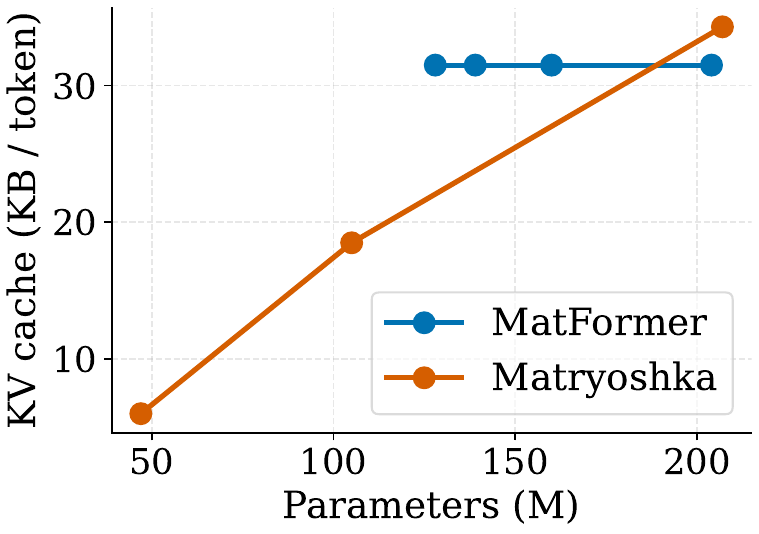}
        \caption{KV cache per token}
        \label{fig:matformer:kv}
    \end{subfigure}
    \caption{MatFormer vs.\ Matryoshka suites at 200M scale, as a function of sub-model
      size. Matryoshka achieves a better size--performance frontier. MatFormer's shared attention backbone fixes its KV cache at
      31.5\,KB/token across all sub-models, whereas Matryoshka's KV cache
      scales with the sub-model.}
    \label{fig:matformer}
\end{figure}

We compare our approach against MatFormer~\citep{devvrit2023matformer} at the 200M
scale, training on 10B tokens, where both methods extract a suite of nested sub-models from a single
training run (\Cref{fig:matformer}). Because MatFormer nests along the FFN width
while sharing a single attention backbone, all of its sub-models carry the same
KV-cache footprint (31.5\,KB/token), independent of their parameter count. In
contrast, Matryoshka nests models along depth, so the KV cache shrinks together with the
sub-model (down to 6.0\,KB/token), yielding a genuinely lighter memory profile at
inference time. We also observe a better quality--size trade-off: at matched
validation perplexity ($\sim$21), Matryoshka-100M matches MatFormer-M (139M) with
roughly half the KV cache and fewer parameters, and Matryoshka-200M attains both the
lowest perplexity (17.92) of all compared
models, including Matformer-XL (19.34).

\subsubsection{Ablations}
\label{sec:exp:ablations}
We ablate three design choices: the inter-model junction (with and without
the norm-rescaling and the fresh embedding), the use of distillation
(with and without), and the allocation of layers across sub-models. Our
ablations are performed on a 200M proxy suite, composed of sub-models of
50M / 100M / 200M, where Matryoshka trades 41\% less training compute for
producing all three sub-models in a single run. Each full run for a 200M
Matryoshka suite costs roughly five days of compute on a single A100 GPU, for 20B training tokens.
\Cref{sec:appendix-distil} reports a separate sweep of the distillation coefficient $\alpha_d$.

\paragraph{Inter-model Junction}
Recall that our recipe concatenates a fresh input embedding with a norm-rescaled
model output. We compare it against two simpler alternatives in \Cref{tab:junction}:
concatenation without norm rescaling, and a zero junction that replaces
the concatenated input embeddings with zeros. Our recipe matches the Vanilla baseline
within 0.02 average PPL; removing the norm match introduces a +0.2 PPL
gap, with a stronger impact on the smaller model; the zero junction is off on all model sizes, with a +0.54 PPL gap over our method. We also report the effect of removing distillation from the recipe, which causes a +0.15 PPL average gap more strongly noticeable on the smaller models. Both components are crucial for the performance of smaller sub-models.

\begin{table}[t]
\centering\small
\begin{tabular}{lrrrr}
\toprule
 & \multicolumn{4}{c}{Validation PPL $\downarrow$} \\
\cmidrule(lr){2-5}
Variant & 50M & 100M & 200M & Average \\
\midrule
Vanilla suite & 26.07 & 18.87 & 15.21 & 20.05 \\
\midrule
Zero padding & 26.98 & 18.83 & 16.01 & 20.61 \\
Input embs. + Distil. & 26.69 & 18.44 & 15.67 & 20.27 \\
Input embs. + Norm & 26.62 & 18.57 & \textbf{15.46} & 20.22 \\
Input embs. + Norm + Distil. (ours) & \textbf{26.16} & \textbf{18.43} & 15.62 & \textbf{20.07} \\
\bottomrule
\end{tabular}
\caption{Method ablation at the 200M scale (validation PPL, lower is
better). Our recipe stays within 0.02 average PPL of the Vanilla baseline;
each simpler junction widens the gap.}
\label{tab:junction}
\end{table}

\begin{figure}[t]
    \centering
    \hfill
    \begin{subcaptionblock}{0.32\linewidth}
        \includegraphics[width=\linewidth]{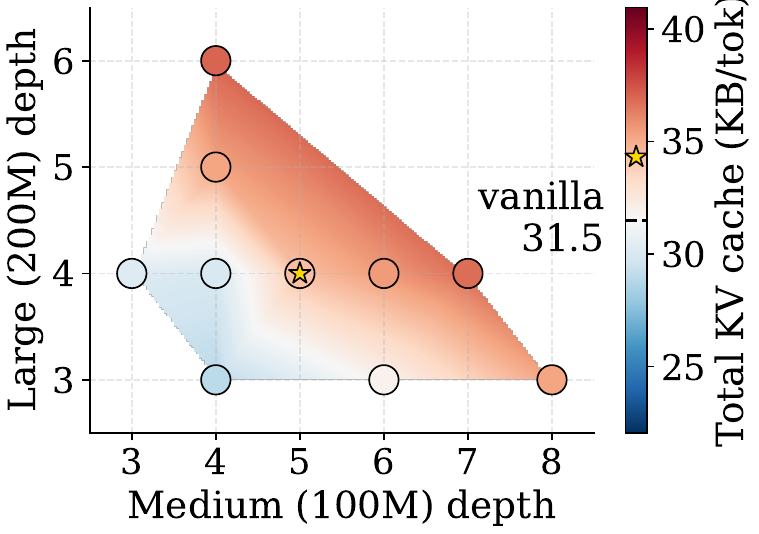}
        \caption{Total KV cache per token}
    \end{subcaptionblock}
    \hfill
    \begin{subcaptionblock}{0.32\linewidth}
        \includegraphics[width=\linewidth]{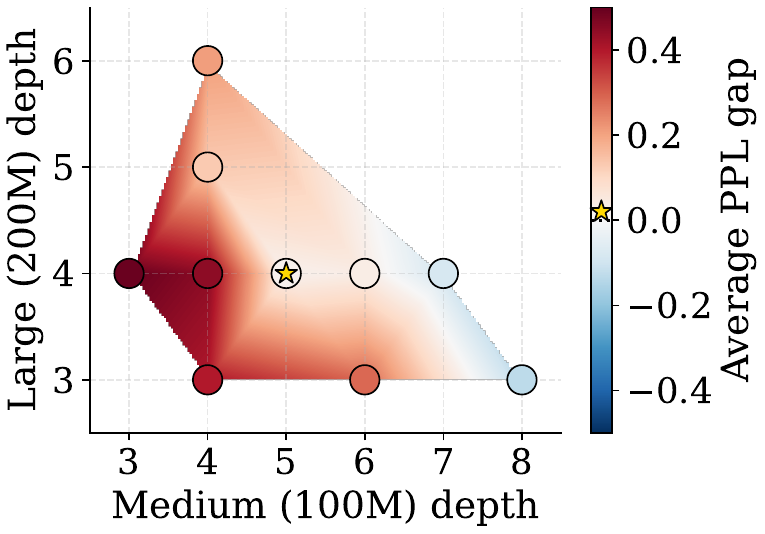}
        \caption{Mean PPL gap}
    \end{subcaptionblock}
    \hfill
    \begin{subcaptionblock}{0.32\linewidth}
        \includegraphics[width=\linewidth]{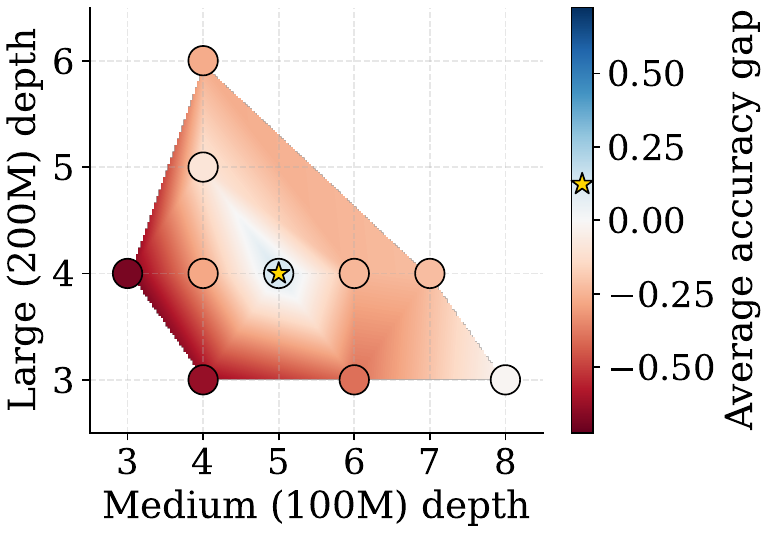}
        \caption{Mean benchmark acc. gap}
    \end{subcaptionblock}
    \hfill\null
    \caption{Shape ablation at the 200M scale. Each point is one
      Matryoshka variant with a shared 50M base sub-model; axes are the
      additional layers of the 100M and 200M blocks. \textbf{(a)} Total KV
      cache per token, with the same-architecture Vanilla 200M baseline
      marked on the colorbar. \textbf{(b)} Mean perplexity gap versus the
      Vanilla baseline, averaged over the three sub-model sizes
      (50M / 100M / 200M); blue means below Vanilla. \textbf{(c)} Mean
      accuracy gap (in percentage points) on the seven downstream
      benchmarks of \Cref{tab:benchmarks}, averaged over the three
      sub-model sizes; blue means above Vanilla.}
    \label{fig:size_ablation}
\end{figure}

\paragraph{Width/Depth of Sub-Models}
The depth triplet of the main 3B suite is one operating point in a rather large design space, as we show in \Cref{fig:sweep_3B}. At 200M parameter scale, the feasible configuration space is much less dense, which allows more coarse exploration. We sweep ten variants at the 200M scale that differ
only in the depth split between the 100M and 200M blocks
(\Cref{fig:size_ablation}). The depth of each sub-model is then fixed to match the parameter size as closely as possible (within 5\% in practice). In \Cref{fig:size_ablation}, we observe that several variants are comparable to the vanilla suite both in terms of total KV cache growth and of average performance gap: the chosen $(L_1, L_2) = (5,4)$ shape (gold star) remains within ${\sim}10\%$ of the KV cache footprint of the Vanilla baseline at very low average PPL cost, and other variants achieve a further
KV reduction with comparable PPL. Additional views (mean
$\Delta\text{NLL}$, per-token FLOPs, 1D KV-cache summary) are reported in
\Cref{sec:appendix-size-shape}. We also select the 200M variant with the
lowest average PPL gap to the Vanilla baseline and report its downstream
benchmark accuracy in \Cref{sec:appendix-200m-bench}, where the same
near-parity pattern as in the 3B suite holds.

\section{Related Work}
\label{sec:related}

\paragraph{Early Exiting}
Early exiting makes predictions at intermediate layers to avoid a full forward
pass when the model is already confident~\citep{teerapittayanon2016branchynet}.
In language modeling, this idea was applied by \citet{schuster2022confident} and
\citet{Elbayad2020Depth-Adaptive}, and more recently via sparse layer routing
by \citet{raposo2024mixtureofdepthsdynamicallyallocatingcompute}.
\citet{elhoushi2024layerskip} combine layer dropout during pretraining with
early-exit inference and self-speculative decoding from the early-exit layers.
All these methods operate within a fixed-width architecture, whereas our
sub-models differ in both width and depth, giving more flexibility in the
accuracy-compute trade-off at each exit point. Furthermore, in Matryoshka suites
there is no ambiguity about where to exit: exits correspond to fully trained
sub-models with their own LM heads, not intermediate classifier probes.

\paragraph{Speculative Decoding and Distillation}
Speculative decoding~\citep{specdec} accelerates autoregressive inference by
letting a small draft model propose candidate continuations that a larger verifier
accepts or rejects in parallel. The key efficiency lever is the token acceptance
rate, which increases when draft and verifier are well aligned.
\citet{zhou2024distillspec} show that distilling the verifier into the draft model
is a principled way to improve alignment.
\citet{zhang-etal-2024-draft} combine speculative decoding with layer skipping for
self-speculative decoding. In the Matryoshka framework, the verifier contains the Transformer stack of the draft model which is a form of static self-speculative decoding, and distillation from the largest sub-model to all smaller ones is a natural by-product of every training step, with negligible additional cost.
More recently, the EAGLE line of work~\citep{li2024eagle,li2024eagle2,li2025eagle3} trains
lightweight feature-level draft heads on top of the verifier's penultimate
representations, achieving 3--5$\times$ speedups.
\citet{cai2024medusa} instead attach multiple parallel decoding heads to a single
model to generate several future tokens at once. These methods are orthogonal to our work and could hypothetically be applied directly to our models and further adapted in our nested framework.
Regarding pretraining-time distillation more broadly,
\citet{peng2025pretrainingdistillation} explore the design space of loss functions
and teacher-student configurations, and \citet{gu2025miniplm} show that offline
teacher logits reduce data requirements by $2.4\times$. Our approach alleviates the need for offline distillation by drastically reducing the cost of online distillation.

\paragraph{Elastic Compression of Pretrained LLMs}
A line of work converts an existing trained LLM into a family of smaller
models via elastic continued training, including
Flextron~\citep{cai2024flextron}, LLaMaFlex~\citep{cai2025llamaflex}, and
Nemotron Elastic~\citep{taghibakhshi2025nemotronelastic}. These methods build
on the supernet lineage of slimmable
networks~\citep{yu2019slimmable,yu2019universallyslimmable} and
Once-for-All~\citep{cai2020onceforall}, and are complementary to our setting:
they recover a family cheaply given a sunk pretraining cost, whereas we
target the from-scratch suite-training problem.

\paragraph{Matryoshka LLMs}
\citet{kusupati2024matryoshkarepresentationlearning} introduce Matryoshka
Representation Learning, where embedding prefixes of any length are trained
to be informative. \citet{devvrit2023matformer} apply the same principle to
Transformers via MatFormer, nesting sub-models by varying the FFN hidden
size across $g$ granularities while keeping depth, attention heads, and
embeddings fixed. This yields FLOPs and parameter savings but leaves the
KV-cache footprint per token identical across all sub-models, since every
granularity inherits the parent's full attention configuration. Sub-models
also are not independently deployable checkpoints. MatFormer thus targets a latency dial within a
single deployment regime, rather than a suite spanning various target regimes with constrained memory usage and detachable per-size checkpoints. Our framework nests both width and depth, and produces standalone checkpoints, at the cost
of a discrete rather than combinatorial set of sizes. \Cref{sec:appendix-matformer}
quantifies this contrast: at a fixed 3B universal shape, MatFormer's
extractable sub-models sit above 1.05B parameters and share a constant
$280$ KB/tok KV cache, whereas Matryoshka tracks the Vanilla curve down
to a $500$M sub-model with $96$ KB/tok of KV cache.
\section{Conclusion}\label{sec:conclusion}

We introduce Matryoshka Language Model Suites, a pretraining framework that
stacks sub-models of increasing size into a single nested architecture trained
end-to-end. The key insight is that the natural structure of a model suite, a
set of models sharing the same training distribution, can be exploited by
nesting at lower cost than independent training. At the 3B scale, our
Matryoshka suite reaches near-parity with independently trained baselines on standard benchmarks while using 36\% less training compute, allowing
low-cost online distillation and enabling speculative decoding with higher
throughput and no memory overhead for the draft model. Our ablations
produce a solid basis for scaling up Matryoshka suites to larger sizes.

Our work paves the way for a new line of research on \emph{joint} pretraining of model suites, where sub-models of different sizes are co-designed rather than trained in isolation. Open questions include how the recipe scales with more sub-models and larger budgets, how to allocate capacity and loss weight across sub-models, and how nested suites interact with post-training stages such as instruction tuning, alignment, and reasoning-oriented finetuning.

\begin{ack}
This work is partially supported by the National Science Foundation NSF under award OAC-2311521 and NASA under award No. 20-OSTFL20-0053.
NG is supported by an Empire AI Postdoctoral Fellowship.
We gratefully acknowledge use of the research
computing resources of the Empire AI Consortium, Inc, with support from the State of New York, the
Simons Foundation, and the Secunda Family Foundation~\citep{10.1145/3708035.3736070}. We would like to gratefully acknowledge the generous support of the NVIDIA Academic Grant Program. This work was supported as part of the “Swiss AI initiative” by a grant from the Swiss National Supercomputing Centre (CSCS) under project ID \texttt{a0137} on Alps.
\end{ack}

\bibliographystyle{plainnat}
\bibliography{local}

@misc{allal2025smollm2smolgoesbig,
      title={SmolLM2: When Smol Goes Big -- Data-Centric Training of a Small Language Model}, 
      author={Loubna Ben Allal and Anton Lozhkov and Elie Bakouch and Gabriel Martín Blázquez and Guilherme Penedo and Lewis Tunstall and Andrés Marafioti and Hynek Kydlíček and Agustín Piqueres Lajarín and Vaibhav Srivastav and Joshua Lochner and Caleb Fahlgren and Xuan-Son Nguyen and Clémentine Fourrier and Ben Burtenshaw and Hugo Larcher and Haojun Zhao and Cyril Zakka and Mathieu Morlon and Colin Raffel and Leandro von Werra and Thomas Wolf},
      year={2025},
      eprint={2502.02737},
      archivePrefix={arXiv},
      primaryClass={cs.CL},
      url={https://arxiv.org/abs/2502.02737}, 
}

@inproceedings{10.1145/3708035.3736070,
author = {Bloom, Stacie and Brumberg, Joshua and Fisk, Ian and Harrison, Robert and Hull, Robert and Ramasubramanian, Melur and Van Vliet, Krystyn and Wing, Jeannette},
title = {Empire AI: A new model for provisioning AI and HPC for academic research in the public good},
year = {2025},
isbn = {9798400713989},
publisher = {Association for Computing Machinery},
address = {New York, NY, USA},
url = {https://doi.org/10.1145/3708035.3736070},
doi = {10.1145/3708035.3736070},
booktitle = {Practice and Experience in Advanced Research Computing 2025: The Power of Collaboration},
articleno = {52},
numpages = {4},
location = {
},
series = {PEARC '25}
}

@article{schuster2022confident,
  title={Confident adaptive language modeling},
  author={Schuster, Tal and Fisch, Adam and Gupta, Jai and Dehghani, Mostafa and Bahri, Dara and Tran, Vinh and Tay, Yi and Metzler, Donald},
  journal={Advances in Neural Information Processing Systems},
  volume={35},
  pages={17456--17472},
  year={2022}
}

@inproceedings{
Elbayad2020Depth-Adaptive,
title={Depth-Adaptive Transformer},
author={Maha Elbayad and Jiatao Gu and Edouard Grave and Michael Auli},
booktitle={International Conference on Learning Representations},
year={2020},
url={https://openreview.net/forum?id=SJg7KhVKPH}
}

@misc{raposo2024mixtureofdepthsdynamicallyallocatingcompute,
      title={Mixture-of-Depths: Dynamically allocating compute in transformer-based language models}, 
      author={David Raposo and Sam Ritter and Blake Richards and Timothy Lillicrap and Peter Conway Humphreys and Adam Santoro},
      year={2024},
      eprint={2404.02258},
      archivePrefix={arXiv},
      primaryClass={cs.LG},
      url={https://arxiv.org/abs/2404.02258}, 
}

@inproceedings{specdec,
author = {Leviathan, Yaniv and Kalman, Matan and Matias, Yossi},
title = {Fast inference from transformers via speculative decoding},
year = {2023},
publisher = {JMLR.org},
booktitle = {Proceedings of the 40th International Conference on Machine Learning},
articleno = {795},
numpages = {13},
location = {Honolulu, Hawaii, USA},
series = {ICML'23}
}

@inproceedings{
zhou2024distillspec,
title={DistillSpec: Improving Speculative Decoding via Knowledge Distillation},
author={Yongchao Zhou and Kaifeng Lyu and Ankit Singh Rawat and Aditya Krishna Menon and Afshin Rostamizadeh and Sanjiv Kumar and Jean-Fran{\c{c}}ois Kagy and Rishabh Agarwal},
booktitle={The Twelfth International Conference on Learning Representations},
year={2024},
url={https://openreview.net/forum?id=rsY6J3ZaTF}
}

@inproceedings{zhang-etal-2024-draft,
    title = "Draft {\&} Verify: Lossless Large Language Model Acceleration via Self-Speculative Decoding",
    author = "Zhang, Jun  and
      Wang, Jue  and
      Li, Huan  and
      Shou, Lidan  and
      Chen, Ke  and
      Chen, Gang  and
      Mehrotra, Sharad",
    editor = "Ku, Lun-Wei  and
      Martins, Andre  and
      Srikumar, Vivek",
    booktitle = "Proceedings of the 62nd Annual Meeting of the Association for Computational Linguistics (Volume 1: Long Papers)",
    month = aug,
    year = "2024",
    address = "Bangkok, Thailand",
    publisher = "Association for Computational Linguistics",
    url = "https://aclanthology.org/2024.acl-long.607/",
    doi = "10.18653/v1/2024.acl-long.607",
    pages = "11263--11282"
}

@InProceedings{paperno2016lambada,
  author    = {Paperno, Denis  and  Kruszewski, Germ\'{a}n  and  Lazaridou,
Angeliki  and  Pham, Ngoc Quan  and  Bernardi, Raffaella  and  Pezzelle,
Sandro  and  Baroni, Marco  and  Boleda, Gemma  and  Fernandez, Raquel},
  title     = {The {LAMBADA} dataset: Word prediction requiring a broad
discourse context},
  booktitle = {Proceedings of the 54th Annual Meeting of the Association for
Computational Linguistics (Volume 1: Long Papers)},
  month     = {August},
  year      = {2016},
  address   = {Berlin, Germany},
  publisher = {Association for Computational Linguistics},
  pages     = {1525--1534},
  url       = {http://www.aclweb.org/anthology/P16-1144}
}

@inproceedings{obqa,
    title = "Can a Suit of Armor Conduct Electricity? A New Dataset for Open Book Question Answering",
    author = "Mihaylov, Todor  and
      Clark, Peter  and
      Khot, Tushar  and
      Sabharwal, Ashish",
    editor = "Riloff, Ellen  and
      Chiang, David  and
      Hockenmaier, Julia  and
      Tsujii, Jun{'}ichi",
    booktitle = "Proceedings of the 2018 Conference on Empirical Methods in Natural Language Processing",
    month = oct # "-" # nov,
    year = "2018",
    address = "Brussels, Belgium",
    publisher = "Association for Computational Linguistics",
    url = "https://aclanthology.org/D18-1260/",
    doi = "10.18653/v1/D18-1260",
    pages = "2381--2391"
}

@inproceedings{piqa,
  author       = {Yonatan Bisk and
                  Rowan Zellers and
                  Ronan Le Bras and
                  Jianfeng Gao and
                  Yejin Choi},
  title        = {{PIQA:} Reasoning about Physical Commonsense in Natural Language},
  booktitle    = {The Thirty-Fourth {AAAI} Conference on Artificial Intelligence, {AAAI}
                  2020, The Thirty-Second Innovative Applications of Artificial Intelligence
                  Conference, {IAAI} 2020, The Tenth {AAAI} Symposium on Educational
                  Advances in Artificial Intelligence, {EAAI} 2020, New York, NY, USA,
                  February 7-12, 2020},
  pages        = {7432--7439},
  publisher    = {{AAAI} Press},
  year         = {2020},
  url          = {https://doi.org/10.1609/aaai.v34i05.6239},
  doi          = {10.1609/AAAI.V34I05.6239},
  bibsource    = {dblp computer science bibliography, https://dblp.org}
}

@article{winogrande,
author = {Sakaguchi, Keisuke and Bras, Ronan Le and Bhagavatula, Chandra and Choi, Yejin},
title = {WinoGrande: an adversarial winograd schema challenge at scale},
year = {2021},
issue_date = {September 2021},
publisher = {Association for Computing Machinery},
address = {New York, NY, USA},
volume = {64},
number = {9},
issn = {0001-0782},
url = {https://doi.org/10.1145/3474381},
doi = {10.1145/3474381},
journal = {Commun. ACM},
month = aug,
pages = {99–106},
numpages = {8}
}

@inproceedings{
w103,
title={Pointer Sentinel Mixture Models},
author={Stephen Merity and Caiming Xiong and James Bradbury and Richard Socher},
booktitle={International Conference on Learning Representations},
year={2017},
url={https://openreview.net/forum?id=Byj72udxe}
}

@inproceedings{c4,
    title = "Documenting Large Webtext Corpora: A Case Study on the Colossal Clean Crawled Corpus",
    author = "Dodge, Jesse  and
      Sap, Maarten  and
      Marasovi{\'c}, Ana  and
      Agnew, William  and
      Ilharco, Gabriel  and
      Groeneveld, Dirk  and
      Mitchell, Margaret  and
      Gardner, Matt",
    editor = "Moens, Marie-Francine  and
      Huang, Xuanjing  and
      Specia, Lucia  and
      Yih, Scott Wen-tau",
    booktitle = "Proceedings of the 2021 Conference on Empirical Methods in Natural Language Processing",
    month = nov,
    year = "2021",
    address = "Online and Punta Cana, Dominican Republic",
    publisher = "Association for Computational Linguistics",
    url = "https://aclanthology.org/2021.emnlp-main.98/",
    doi = "10.18653/v1/2021.emnlp-main.98",
    pages = "1286--1305"
}

@inproceedings{
pg19,
title={Compressive Transformers for Long-Range Sequence Modelling},
author={Jack W. Rae and Anna Potapenko and Siddhant M. Jayakumar and Chloe Hillier and Timothy P. Lillicrap},
booktitle={International Conference on Learning Representations},
year={2020},
url={https://openreview.net/forum?id=SylKikSYDH}
}

@inproceedings{
cai2024medusa,
title={Medusa: Simple {LLM} Inference Acceleration Framework with Multiple Decoding Heads},
author={Tianle Cai and Yuhong Li and Zhengyang Geng and Hongwu Peng and Jason D. Lee and Deming Chen and Tri Dao},
booktitle={Forty-first International Conference on Machine Learning},
year={2024},
url={https://openreview.net/forum?id=PEpbUobfJv}
}

@inproceedings{
devvrit2023matformer,
title={MatFormer: Nested Transformer for Elastic Inference},
author={Fnu Devvrit and Sneha Kudugunta and Aditya Kusupati and Tim Dettmers and Kaifeng Chen and Inderjit Dhillon and Yulia Tsvetkov and Hannaneh Hajishirzi and Sham Kakade and Ali Farhadi and Prateek Jain},
booktitle={Workshop on Advancing Neural Network Training: Computational Efficiency, Scalability, and Resource Optimization (WANT@NeurIPS 2023)},
year={2023},
url={https://openreview.net/forum?id=93BaEweoRg}
}

@inproceedings{kusupati2024matryoshkarepresentationlearning,
  title     = {Matryoshka Representation Learning},
  author    = {Kusupati, Aditya and Bhatt, Gantavya and Rege, Aniket and Wallingford, Matthew and Sinha, Aditya and Ramanujan, Vivek and Howard-Snyder, William and Chen, Kaifeng and Kakade, Sham and Jain, Prateek and others},
  title     = {Matryoshka Representation Learning.},
  booktitle = {Advances in Neural Information Processing Systems},
  month     = {December},
  year      = {2022},
}

@inproceedings{NIPS2017_3f5ee243,
 author = {Vaswani, Ashish and Shazeer, Noam and Parmar, Niki and Uszkoreit, Jakob and Jones, Llion and Gomez, Aidan N and Kaiser, \L ukasz and Polosukhin, Illia},
 booktitle = {Advances in Neural Information Processing Systems},
 editor = {I. Guyon and U. Von Luxburg and S. Bengio and H. Wallach and R. Fergus and S. Vishwanathan and R. Garnett},
 pages = {},
 publisher = {Curran Associates, Inc.},
 title = {Attention is All you Need},
 url = {https://proceedings.neurips.cc/paper_files/paper/2017/file/3f5ee243547dee91fbd053c1c4a845aa-Paper.pdf},
 volume = {30},
 year = {2017}
}

@article{llama3modelcard,
title={Llama 3 Model Card},
author={AI@Meta},
year={2024},
url = {https://github.com/meta-llama/llama3/blob/main/MODEL_CARD.md}
}

@misc{allenai:arc,
      title={Think you have Solved Question Answering? Try ARC, the AI2 Reasoning Challenge}, 
      author={Peter Clark and Isaac Cowhey and Oren Etzioni and Tushar Khot and Ashish Sabharwal and Carissa Schoenick and Oyvind Tafjord},
      year={2018},
      eprint={1803.05457},
      archivePrefix={arXiv},
      primaryClass={cs.AI},
      url={https://arxiv.org/abs/1803.05457}, 
}

@article{rope,
author = {Su, Jianlin and Ahmed, Murtadha and Lu, Yu and Pan, Shengfeng and Bo, Wen and Liu, Yunfeng},
title = {RoFormer: Enhanced transformer with Rotary Position Embedding},
year = {2024},
issue_date = {Feb 2024},
publisher = {Elsevier Science Publishers B. V.},
address = {NLD},
volume = {568},
number = {C},
issn = {0925-2312},
url = {https://doi.org/10.1016/j.neucom.2023.127063},
doi = {10.1016/j.neucom.2023.127063},
journal = {Neurocomput.},
month = feb,
numpages = {12}
}

@inproceedings{zellers2019hellaswag,
    title = "{H}ella{S}wag: Can a Machine Really Finish Your Sentence?",
    author = "Zellers, Rowan  and
      Holtzman, Ari  and
      Bisk, Yonatan  and
      Farhadi, Ali  and
      Choi, Yejin",
    editor = "Korhonen, Anna  and
      Traum, David  and
      M{\`a}rquez, Llu{\'i}s",
    booktitle = "Proceedings of the 57th Annual Meeting of the Association for Computational Linguistics",
    month = jul,
    year = "2019",
    address = "Florence, Italy",
    publisher = "Association for Computational Linguistics",
    url = "https://aclanthology.org/P19-1472/",
    doi = "10.18653/v1/P19-1472",
    pages = "4791--4800"
}

@inproceedings{
  penedo2024the,
  title={The FineWeb Datasets: Decanting the Web for the Finest Text Data at Scale},
  author={Guilherme Penedo and Hynek Kydl{\'\i}{\v{c}}ek and Loubna Ben allal and Anton Lozhkov and Margaret Mitchell and Colin Raffel and Leandro Von Werra and Thomas Wolf},
  booktitle={The Thirty-eight Conference on Neural Information Processing Systems Datasets and Benchmarks Track},
  year={2024},
  url={https://openreview.net/forum?id=n6SCkn2QaG}
}

@misc{deepseekai2026deepseekv4,
      title={DeepSeek-V4: Towards Highly Efficient Million-Token Context Intelligence},
      author={DeepSeek-AI},
      url={https://huggingface.co/deepseek-ai/DeepSeek-V4-Pro/blob/main/DeepSeek_V4.pdf},
      year={2026},
}

@inproceedings{elhoushi2024layerskip,
    title = "{L}ayer{S}kip: Enabling Early Exit Inference and Self-Speculative Decoding",
    author = "Elhoushi, Mostafa  and
      Shrivastava, Akshat  and
      Liskovich, Diana  and
      Hosmer, Basil  and
      Wasti, Bram  and
      Lai, Liangzhen  and
      Mahmoud, Anas  and
      Acun, Bilge  and
      Agarwal, Saurabh  and
      Roman, Ahmed  and
      Aly, Ahmed  and
      Chen, Beidi  and
      Wu, Carole-Jean",
    editor = "Ku, Lun-Wei  and
      Martins, Andre  and
      Srikumar, Vivek",
    booktitle = "Proceedings of the 62nd Annual Meeting of the Association for Computational Linguistics (Volume 1: Long Papers)",
    month = aug,
    year = "2024",
    address = "Bangkok, Thailand",
    publisher = "Association for Computational Linguistics",
    url = "https://aclanthology.org/2024.acl-long.681/",
    doi = "10.18653/v1/2024.acl-long.681",
    pages = "12622--12642"
}

@inproceedings{li2024eagle,
author = {Li, Yuhui and Wei, Fangyun and Zhang, Chao and Zhang, Hongyang},
title = {{EAGLE}: speculative sampling requires rethinking feature uncertainty},
year = {2024},
publisher = {JMLR.org},
booktitle = {Proceedings of the 41st International Conference on Machine Learning},
articleno = {1162},
numpages = {14},
location = {Vienna, Austria},
series = {ICML'24}
}

@inproceedings{li2024eagle2,
    title = "{EAGLE}-2: Faster Inference of Language Models with Dynamic Draft Trees",
    author = "Li, Yuhui  and
      Wei, Fangyun  and
      Zhang, Chao  and
      Zhang, Hongyang",
    editor = "Al-Onaizan, Yaser  and
      Bansal, Mohit  and
      Chen, Yun-Nung",
    booktitle = "Proceedings of the 2024 Conference on Empirical Methods in Natural Language Processing",
    month = nov,
    year = "2024",
    address = "Miami, Florida, USA",
    publisher = "Association for Computational Linguistics",
    url = "https://aclanthology.org/2024.emnlp-main.422/",
    doi = "10.18653/v1/2024.emnlp-main.422",
    pages = "7421--7432"
}

@inproceedings{
li2025eagle3,
title={{EAGLE}-3: Scaling up Inference Acceleration of Large Language Models via Training-Time Test},
author={Yuhui Li and Fangyun Wei and Chao Zhang and Hongyang Zhang},
booktitle={The Thirty-ninth Annual Conference on Neural Information Processing Systems},
year={2026},
url={https://openreview.net/forum?id=4exx1hUffq}
}

@inproceedings{peng2025pretrainingdistillation,
    title = "Pre-training Distillation for Large Language Models: A Design Space Exploration",
    author = "Peng, Hao  and
      Lv, Xin  and
      Bai, Yushi  and
      Yao, Zijun  and
      Zhang, Jiajie  and
      Hou, Lei  and
      Li, Juanzi",
    editor = "Che, Wanxiang  and
      Nabende, Joyce  and
      Shutova, Ekaterina  and
      Pilehvar, Mohammad Taher",
    booktitle = "Proceedings of the 63rd Annual Meeting of the Association for Computational Linguistics (Volume 1: Long Papers)",
    month = jul,
    year = "2025",
    address = "Vienna, Austria",
    publisher = "Association for Computational Linguistics",
    url = "https://aclanthology.org/2025.acl-long.181/",
    doi = "10.18653/v1/2025.acl-long.181",
    pages = "3603--3618",
    ISBN = "979-8-89176-251-0"
}

@inproceedings{
gu2025miniplm,
title={Mini{PLM}: Knowledge Distillation for Pre-training Language Models},
author={Yuxian Gu and Hao Zhou and Fandong Meng and Jie Zhou and Minlie Huang},
booktitle={The Thirteenth International Conference on Learning Representations},
year={2025},
url={https://openreview.net/forum?id=tJHDw8XfeC}
}

@article{teerapittayanon2016branchynet,
  title={BranchyNet: Fast inference via early exiting from deep neural networks},
  author={Surat Teerapittayanon and Bradley McDanel and H. T. Kung},
  journal={2016 23rd International Conference on Pattern Recognition (ICPR)},
  year={2016},
  pages={2464-2469},
  url={https://api.semanticscholar.org/CorpusID:2916466}
}

@inproceedings{
yu2019slimmable,
title={Slimmable Neural Networks},
author={Jiahui Yu and Linjie Yang and Ning Xu and Jianchao Yang and Thomas Huang},
booktitle={International Conference on Learning Representations},
year={2019},
url={https://openreview.net/forum?id=H1gMCsAqY7},
}

@INPROCEEDINGS {yu2019universallyslimmable,
author = { Yu, Jiahui and Huang, Thomas },
booktitle = { 2019 IEEE/CVF International Conference on Computer Vision (ICCV) },
title = {{ Universally Slimmable Networks and Improved Training Techniques }},
year = {2019},
volume = {},
ISSN = {},
pages = {1803-1811},
doi = {10.1109/ICCV.2019.00189},
url = {https://doi.ieeecomputersociety.org/10.1109/ICCV.2019.00189},
publisher = {IEEE Computer Society},
address = {Los Alamitos, CA, USA},
month =Nov}

@inproceedings{
cai2020onceforall,
title={Once-for-All: Train One Network and Specialize it for Efficient Deployment},
author={Han Cai and Chuang Gan and Tianzhe Wang and Zhekai Zhang and Song Han},
booktitle={International Conference on Learning Representations},
year={2020},
url={https://openreview.net/forum?id=HylxE1HKwS}
}

@inproceedings{
        cai2024flextron,
        title={Flextron: Many-in-One Flexible Large Language Model},
        author={Ruisi Cai and Saurav Muralidharan and Greg Heinrich and Hongxu Yin and Zhangyang Wang and Jan Kautz and Pavlo Molchanov},
        booktitle={Forty-first International Conference on Machine Learning},
        year={2024},
        url={https://openreview.net/forum?id=9vKRhnflAs}
        }

@inproceedings{
cai2025llamaflex,
title={{LL}aMaFlex: Many-in-one {LLM}s via Generalized Pruning and Weight Sharing},
author={Ruisi Cai and Saurav Muralidharan and Hongxu Yin and Zhangyang Wang and Jan Kautz and Pavlo Molchanov},
booktitle={The Thirteenth International Conference on Learning Representations},
year={2025},
url={https://openreview.net/forum?id=AyC4uxx2HW}
}

@misc{taghibakhshi2025nemotronelastic,
      title={Nemotron Elastic: Towards Efficient Many-in-One Reasoning LLMs}, 
      author={Ali Taghibakhshi and Sharath Turuvekere Sreenivas and Saurav Muralidharan and Ruisi Cai and Marcin Chochowski and Ameya Sunil Mahabaleshwarkar and Yoshi Suhara and Oluwatobi Olabiyi and Daniel Korzekwa and Mostofa Patwary and Mohammad Shoeybi and Jan Kautz and Bryan Catanzaro and Ashwath Aithal and Nima Tajbakhsh and Pavlo Molchanov},
      year={2025},
      eprint={2511.16664},
      archivePrefix={arXiv},
      primaryClass={cs.CL},
      url={https://arxiv.org/abs/2511.16664}, 
}

@misc{kaplan2020scalinglaws,
      title={Scaling Laws for Neural Language Models}, 
      author={Jared Kaplan and Sam McCandlish and Tom Henighan and Tom B. Brown and Benjamin Chess and Rewon Child and Scott Gray and Alec Radford and Jeffrey Wu and Dario Amodei},
      year={2020},
      eprint={2001.08361},
      archivePrefix={arXiv},
      primaryClass={cs.LG},
      url={https://arxiv.org/abs/2001.08361}, 
}

@inproceedings{tay2022scalingarchitectures,
    title = "Scaling Laws vs Model Architectures: How does Inductive Bias Influence Scaling?",
    author = "Tay, Yi  and
      Dehghani, Mostafa  and
      Abnar, Samira  and
      Chung, Hyung  and
      Fedus, William  and
      Rao, Jinfeng  and
      Narang, Sharan  and
      Tran, Vinh  and
      Yogatama, Dani  and
      Metzler, Donald",
    editor = "Bouamor, Houda  and
      Pino, Juan  and
      Bali, Kalika",
    booktitle = "Findings of the Association for Computational Linguistics: EMNLP 2023",
    month = dec,
    year = "2023",
    address = "Singapore",
    publisher = "Association for Computational Linguistics",
    url = "https://aclanthology.org/2023.findings-emnlp.825/",
    doi = "10.18653/v1/2023.findings-emnlp.825",
    pages = "12342--12364"
}

@inproceedings{petty2024depth,
    title = "The Impact of Depth on Compositional Generalization in Transformer Language Models",
    author = "Petty, Jackson  and
      Steenkiste, Sjoerd  and
      Dasgupta, Ishita  and
      Sha, Fei  and
      Garrette, Dan  and
      Linzen, Tal",
    editor = "Duh, Kevin  and
      Gomez, Helena  and
      Bethard, Steven",
    booktitle = "Proceedings of the 2024 Conference of the North American Chapter of the Association for Computational Linguistics: Human Language Technologies (Volume 1: Long Papers)",
    month = jun,
    year = "2024",
    address = "Mexico City, Mexico",
    publisher = "Association for Computational Linguistics",
    url = "https://aclanthology.org/2024.naacl-long.402/",
    doi = "10.18653/v1/2024.naacl-long.402",
    pages = "7239--7252"
}

\appendix
\crefalias{section}{appendix}
\section{Distillation Ablation}
\label{sec:appendix-distil}

\begin{figure}[h]
    \centering
    \includegraphics[height=1.2em]{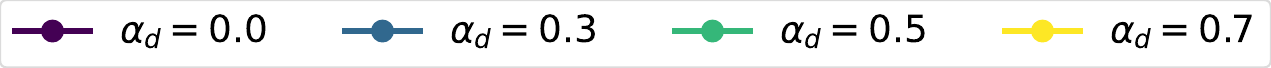}\hspace{1em}\includegraphics[height=1.2em]{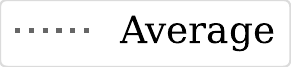}

    \hfill
    \begin{subcaptionblock}{0.4\linewidth}
        \includegraphics[width=\linewidth]{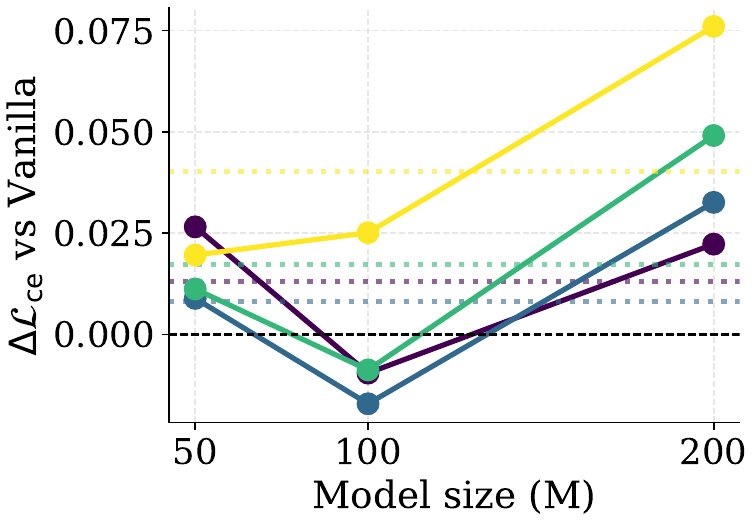}
        \caption{Validation NLL gap vs Vanilla}
    \end{subcaptionblock}
    \hfill
    \begin{subcaptionblock}{0.4\linewidth}
        \includegraphics[width=\linewidth]{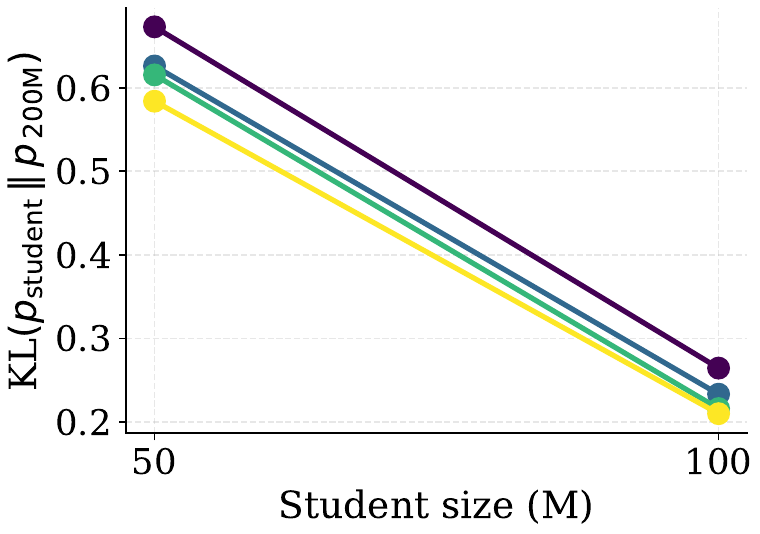}
        \caption{KL$(p_{\text{student}} \,\|\, p_{\,200\text{M}})$}
    \end{subcaptionblock}
    \hfill\null

    \caption{Effect of the distillation coefficient $\alpha_d$ on the 200M
      proxy suite. \textbf{(a)} Validation cross-entropy of each Matryoshka
      sub-model minus its Vanilla counterpart; dotted lines are the average
      across sub-model sizes. $\alpha_d = 0.3$ minimizes the average gap;
      higher values harm the larger sub-models. \textbf{(b)} KL divergence
      of each student sub-model (50M, 100M) to the 200M teacher, with one
      curve per $\alpha_d$ (color); KL drops monotonically as $\alpha_d$
      increases at both student sizes.}
    \label{fig:distil}
\end{figure}

We sweep the distillation coefficient $\alpha_d \in \{0, 0.3, 0.5, 0.7\}$
on the 200M proxy suite (sub-models 50M / 100M / 200M, otherwise identical
recipe to the main runs). \Cref{fig:distil}(a) reports the validation cross-entropy of
each Matryoshka sub-model relative to its Vanilla-baseline counterpart;
\Cref{fig:distil}(b) reports the KL divergence of each smaller sub-model
to the 200M teacher, which captures how aligned each student becomes with
the teacher under each setting.

In \Cref{fig:distil}(a), $\alpha_d = 0$ (no distillation) leaves the
50M and 200M sub-models above Vanilla while the intermediate 100M
sub-model already sits below it; turning on the distillation signal
($\alpha_d = 0.3$) closes most of the 50M gap and pushes the average gap
to its minimum across the four settings. Pushing $\alpha_d$ higher is
counter-productive: at $\alpha_d = 0.5$ the largest sub-model starts
trailing Vanilla, and $\alpha_d = 0.7$ degrades every size. The pattern
in \Cref{fig:distil}(b) mirrors this — KL drops monotonically with
$\alpha_d$, confirming that larger coefficients pull the student
distribution closer to the teacher's, but at a cost on absolute language
modeling quality once the cross-entropy term is too downweighted. We
therefore use $\alpha_d = 0.3$ for the main 3B suite.

\section{Compute, Memory, and Training-FLOPs Accounting}
\label{sec:appendix-accounting}

This appendix details the formulas used throughout the paper to report
parameter counts, KV cache footprint, theoretical inference FLOPs, and
training FLOPs. We use the following notation: $V$ is the vocabulary
size, $L$ the number of Transformer layers, $H$ the hidden dimension
(equal to the product of the number of attention heads and the head
dimension), and $D$ the per-token training dataset size.

\paragraph{Parameter count}
For a standalone Transformer with input embedding, $L$ blocks of width $D$
with $4D$ FFN intermediate dimension and an LM head, the total parameter
count is
\begin{equation}
  N_{\text{std}}(L, D) = 2 V D + 16 L D^2,
  \label{eq:appx-params-std}
\end{equation}
where $2VD$ accounts for the input embedding and the LM head and
$16 L D^2$ accounts for the per-layer cost (4$D^2$ for self-attention
projections and 12$D^2$ for the FFN). For a Matryoshka suite of $M$
nested sub-models with shapes $(n_m, D_m)_{m=1}^M$, $D_1 \leq \cdots \leq  D_M$,
the cumulative parameter count observed at sub-model $m$ is
\begin{equation}
  N^{\text{cumul}}_m = 2 V D_m + \sum_{j \le m} 16 \, n_j D_j^2,
  \label{eq:appx-params-mat}
\end{equation}
where the $2 V D_m$ term reflects that sub-model $m$ uses the full input
embedding at width $D_m$ together with its own LM head $W^m_\theta$. The
total trainable parameter count of the suite further adds the LM heads of
the smaller sub-models, $\sum_{m < M} V D_m$.

\paragraph{KV Cache}
For a single decoded token at full attention, each Transformer layer
stores both keys and values of width $D$, giving
\begin{equation}
  \text{KV}(L, D) = 2 \, b \, L \, D \quad \text{bytes per token},
  \label{eq:appx-kv}
\end{equation}
with $b$ bytes per element ($b = 2$ for bf16).

\paragraph{Theoretical Inference FLOPs}
Counting two FLOPs per multiply-accumulate, the per-token forward-pass
FLOPs of a standalone Transformer stack of width $D$ and depth $L$ is
\begin{equation}
  F_{\text{stack}}(L, D) = 2 \cdot 16 \, L D^2 = 32 \, L D^2,
  \label{eq:appx-flops-stack}
\end{equation}
to which we add a single LM head application of $2 V H$ FLOPs to produce
the next-token logits. For the full Matryoshka suite, the per-token
inference cost is the sum of the stack costs across all sub-models plus
a single LM head at the deepest exit:
\begin{equation}
  F^{\text{Mat}}_{\text{infer}} = \sum_{m=1}^M F_{\text{stack}}(n_m, D_m)
    + 2 V D_M.
  \label{eq:appx-flops-mat}
\end{equation}
The values reported in \Cref{fig:sweep_3B} and elsewhere use these
formulas directly.

\paragraph{Training FLOPs}
We report training FLOPs by combining wall-clock GPU-hours with the
nominal BF16 Tensor-Core peak of the GPU used for the run:
\begin{equation}
  F_{\text{train}} = \frac{D}{(\text{tokens/GPU-hour})}
    \times 3600 \times F_{\text{peak}}^{\text{bf16}},
  \label{eq:appx-train-flops}
\end{equation}
where $D$ is the number of training tokens, the tokens-per-GPU-hour
throughput is measured from training logs, and $F_{\text{peak}}^{\text{bf16}}$
is the GPU's nominal BF16 throughput (we use $4.5 \times 10^{15}$ FLOP/s
for an NVIDIA B200 and $3.12 \times 10^{14}$ FLOP/s for an NVIDIA A100).
This convention conflates the model's algorithmic FLOPs with hardware
utilization, but it is the most faithful comparison between Vanilla and
Matryoshka training costs, as any kernel-level differences are reflected in
the measured throughput. Per-suite training FLOPs are the
sum of $F_{\text{train}}$ over the runs that compose the suite (one
joint run for Matryoshka, three independent runs for Vanilla).

\section{Choice of Total Layer Budget}
\label{sec:appendix-layer-budget}

The Matryoshka suite assigns a depth $n_i$ to each sub-model under a fixed
total budget $L = n_1 + n_2 + n_3$. Before searching the depth allocation
inside $L$ (\Cref{fig:sweep_3B}), we first fix $L$ itself. The choice is
constrained by two architectural costs of the 3B sub-model: per-token
FLOPs and total KV cache. The Matryoshka 3B sub-model uses every layer
across the three blocks, so $L$ directly determines its KV-cache footprint
and, together with the per-block widths, its FLOPs.

\begin{figure}[h]
    \centering
    \includegraphics[width=0.7\linewidth]{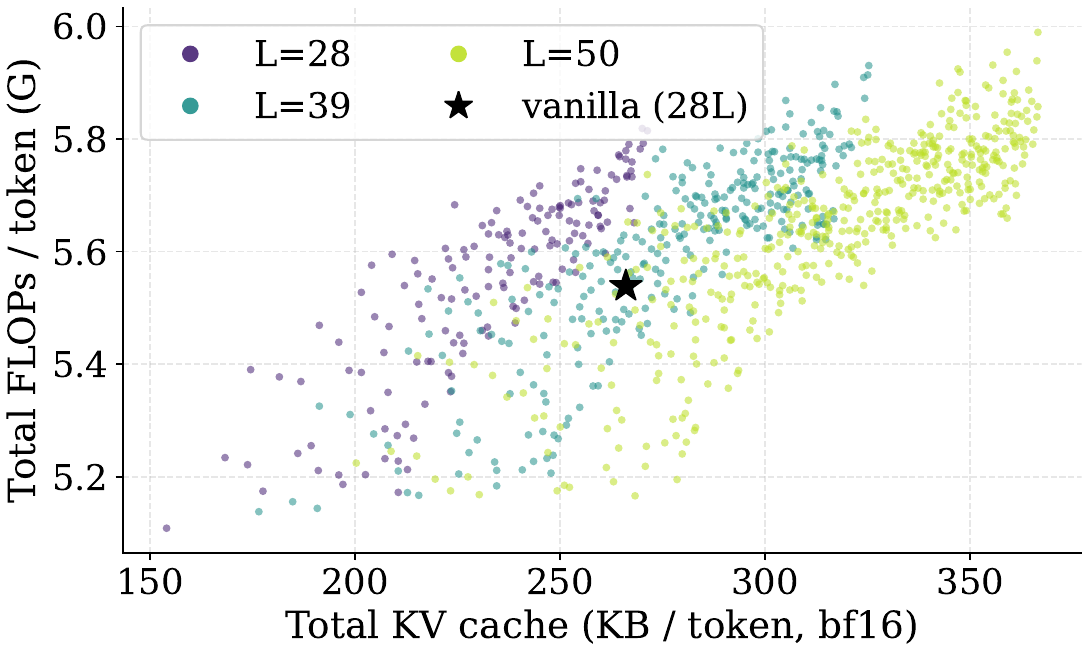}
    \caption{Per-token FLOPs and total KV cache of the Matryoshka 3B
      sub-model across layer budgets $L \in \{28, 39, 50\}$. Each colored
      point is one valid depth allocation $(n_1, n_2, n_3)$ summing to $L$,
      with per-block widths solved to hit the (500M, 1.5B, 3B) parameter
      targets. The black star is the Vanilla 3B baseline (28 layers,
      $D_m = 2\,560$).}
    \label{fig:layer_budget}
\end{figure}

To make the trade-off concrete, we sweep $L \in \{28, 39, 50\}$. For each
$L$, we randomly sample roughly 500 valid depth allocations $(n_1, n_2, n_3)$
summing to $L$, and for each allocation we solve for the per-block widths
that hit the (500M, 1.5B, 3B) parameter targets at fixed head dimensions
$(64, 96, 128)$. We compute the resulting per-token FLOPs and total KV
cache (in bf16) of the full 3B forward pass. \Cref{fig:layer_budget} plots
each sample as a point colored by $L$; the black star marks the Vanilla 3B
baseline (28 layers, $D_m = 2\,560$).

The three clouds occupy overlapping but distinct regions:
\textbf{$L = 28$} clusters at the lowest KV cache, but its FLOPs cloud sits
mostly above the Vanilla baseline, because matching the 3B parameter
target with only 28 shared layers forces wider matrices and higher FLOPs;
\textbf{$L = 50$} reaches the lowest FLOPs but pays a steep KV-cache cost
(${\sim}25{-}40\%$ above Vanilla), since KV cache scales linearly with
depth and is unaffected by the FLOPs-saving width reduction;
\textbf{$L = 39$} contains allocations that simultaneously match or beat
Vanilla on both axes — the only budget for which a strictly Pareto-dominant
configuration exists. We therefore fix $L = 39$ as the depth budget and
search inside it for the allocation $(n_1, n_2, n_3)$, which is the
analysis reported in \Cref{fig:sweep_3B}.

Note that the impact of the width-depth ratio on LM quality is also
studied in the
literature~\citep{kaplan2020scalinglaws,tay2022scalingarchitectures,petty2024depth},
with the general finding that scale matters more than shape over a
reasonable range of ratios. In the case of the Matryoshka models, matching the KV cache footprint matters for matching performance with a standard counterpart (\Cref{sec:appendix-size-shape}).

\section{Reading the Ternary Plots}
\label{sec:appendix-ternary}
\begin{figure}[h]
    \centering
    \includegraphics[width=0.45\linewidth]{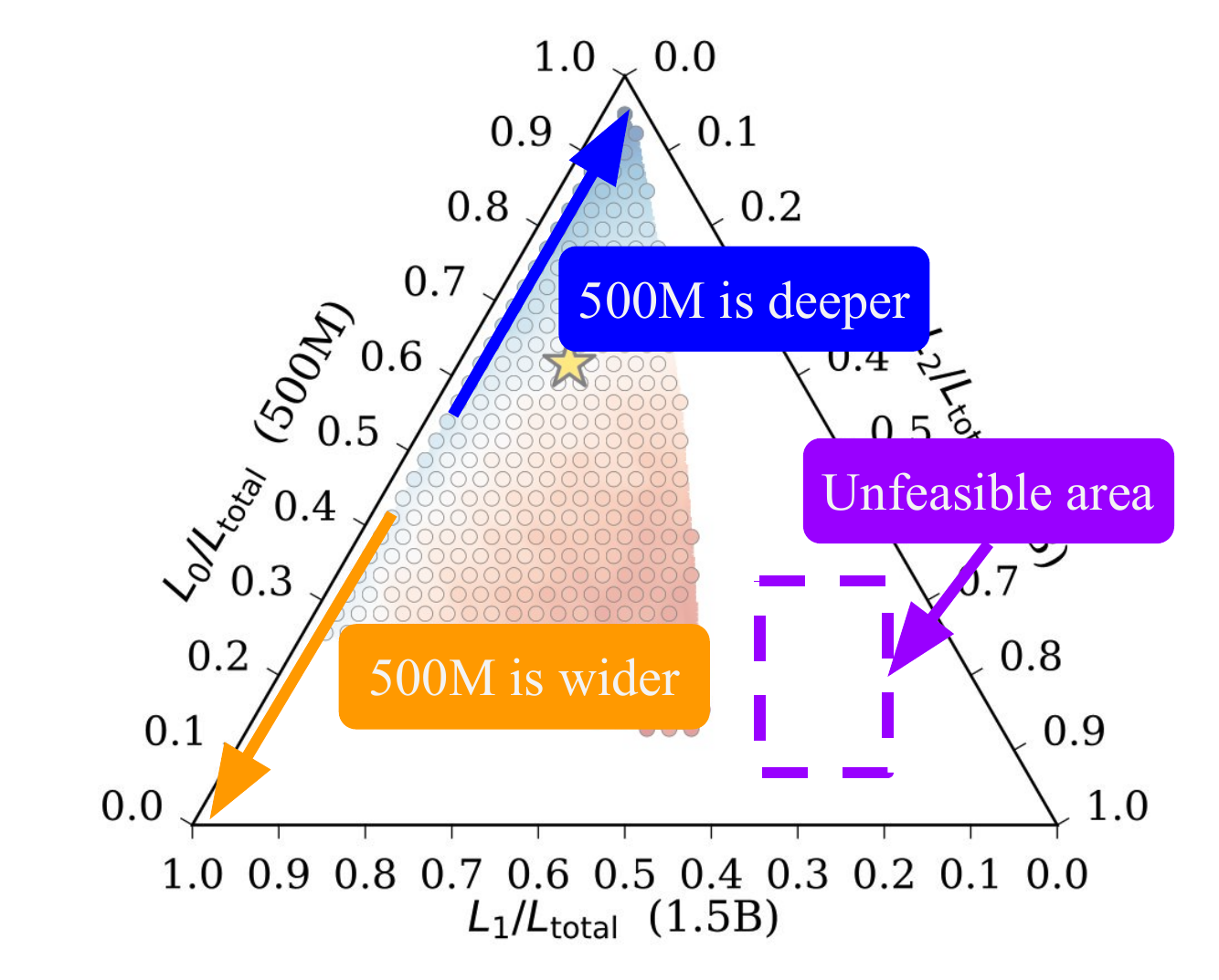}
    \caption{Reading guide for the ternary depth-allocation plots used
      throughout the paper. Vertices, edges, direction of motion, and the
      unfeasible region are annotated; the gold star marks the chosen
      configuration $(24, 10, 5)$.}
    \label{fig:ternary_explain}
\end{figure}
Several figures in this paper visualize the depth allocation $(L_0, L_1,
L_2)$ across the three sub-models on a ternary diagram, with $L_0 + L_1 +
L_2 = L_{\mathrm{total}}$ fixed. Each axis carries the \emph{fraction} of
total layers assigned to one sub-model, $L_i / L_{\mathrm{total}}$, so
every point in the triangle corresponds to one valid normalized depth
triplet. The three vertices represent the degenerate allocations where a
single sub-model receives all layers; the three edges represent allocations
where one sub-model receives zero layers. \Cref{fig:ternary_explain}
illustrates the conventions on the (500M, 1.5B, 3B) sweep at
$L_{\mathrm{total}} = 39$.

Direction of motion has a concrete architectural meaning. Moving a point
toward the top vertex increases $L_0 / L_{\mathrm{total}}$, the fraction
of layers used by the 500M sub-model. To keep the 500M parameter target
fixed, the per-layer width must shrink, so points near the top vertex
correspond to a 500M sub-model that is \emph{deeper and narrower}. Moving
toward the bottom-left vertex decreases $L_0 / L_{\mathrm{total}}$ in
favor of $L_1 / L_{\mathrm{total}}$ (the 1.5B share); the 500M sub-model
then has fewer layers and must be \emph{wider} to hit the same parameter
count. The same logic applies to the other two axes. The gold star marks
the configuration used in our main experiments,
$(L_0, L_1, L_2) = (24, 10, 5)$.

The dashed region marked ``unfeasible'' contains depth triplets for which
no valid Matryoshka architecture exists at the prescribed parameter
targets. Concretely, given the head dimensions
$(d_h^{(0)}, d_h^{(1)}, d_h^{(2)}) = (64, 96, 128)$ and target sizes
(500M, 1.5B, 3B), we solve for the head count of each sub-model so that
the cumulative parameter count matches the target. Two constraints can
fail. Solving for the head count of a sub-model with too few layers can
require fewer than one head, which is not realizable. More commonly, the
nesting requirement $D_0 \leq  D_1 \leq  D_2$ on the hidden dimensions can be
violated: when a larger sub-model is assigned too many layers, the
only way to hit its parameter target is for it to be \emph{narrower} than
its predecessor, which is incompatible with the constraint. In both situations, this will lead to a missing point on the ternary graph.

\section{Resource Comparison with MatFormer}
\label{sec:appendix-matformer}

To clarify how Matryoshka relates to other nested-architecture
proposals, we compare it directly to MatFormer~\citep{devvrit2023matformer}
on the two resource axes that matter at deployment time: KV cache
footprint and per-token inference FLOPs. MatFormer applies its nesting
\emph{only on the FFN block}, leaving attention shape, hidden width, and
depth fixed at the universal-model values. Sub-models are extracted by
truncating the FFN intermediate dimension to a fraction $r \in (0, 4]$ of
the universal $4 H$ width, with canonical granularities at $r \in \{0.5,
1, 2, 4\}$ in the original paper. Mix'n'Match supports any positive ratio
in principle, so we sweep $r$ continuously below.

\begin{figure}[h]
    \centering
    \includegraphics[height=1.2em]{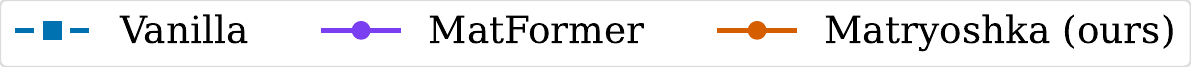}

    \begin{subcaptionblock}{0.45\linewidth}
        \includegraphics[width=\linewidth]{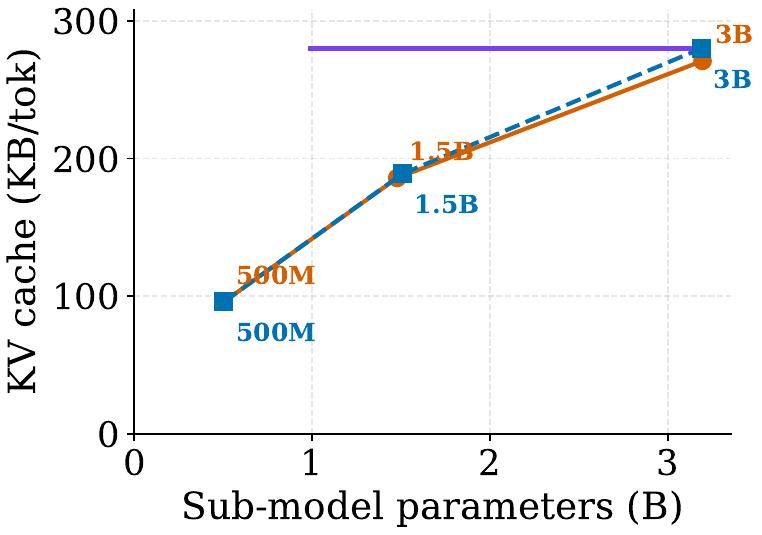}
        \caption{KV cache per token}
        \label{fig:matformer-kv}
    \end{subcaptionblock}
    \hfill
    \begin{subcaptionblock}{0.45\linewidth}
        \includegraphics[width=\linewidth]{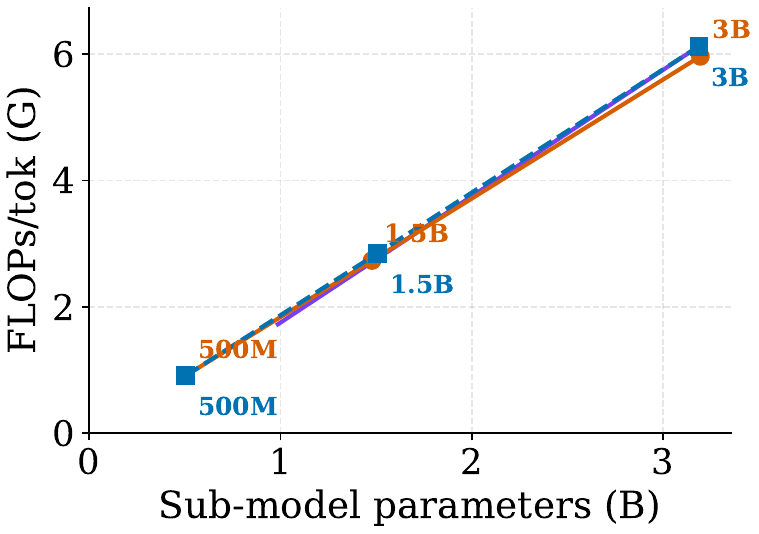}
        \caption{Theoretical inference FLOPs per token}
        \label{fig:matformer-flops}
    \end{subcaptionblock}
    \caption{Resource cost of extracting a sub-model from each framework,
      with the universal shape fixed to the Vanilla 3B baseline ($H =
      2560$, $L = 28$). The MatFormer curve is a continuous sweep of FFN
      ratio $r \in (0, 4]$. Matryoshka and Vanilla are the three actual
      sub-models / independent baselines from \Cref{tab:arch}. Lower-left
      is better on both panels.}
    \label{fig:matformer-comparison}
\end{figure}

We fix the universal-model shape to the Vanilla 3B baseline
(\Cref{tab:arch}, $H = 2560$, $L = 28$) for both methods. For MatFormer we
apply the standalone-Transformer formulas of \Cref{sec:appendix-accounting}
with FFN cost $3 r H^2$ per layer (three SwiGLU
projections~\citep{llama3modelcard} of width $r H$). For Matryoshka we use
the cumulative formulas at the three exit points (500M, 1.5B, 3B) of the
3B suite. \Cref{fig:matformer-comparison} reports the result.

\paragraph{Smallest Extractable Model Floor} The MatFormer trace in
\Cref{fig:matformer-comparison} starts at roughly $1.26$B parameters even
at the smallest canonical granularity $r = 0.5$, and a hypothetical
$r \to 0$ would still leave a $\sim$1.05B floor coming from the input
embedding, the LM head, and the four attention projections at every
layer, which are all kept at the universal width. This means a 3B
universal MatFormer cannot expose a sub-500M extractable model at all,
in either parameter count or FLOPs/tok: \Cref{fig:matformer-flops} shows
that MatFormer bottoms out at $2.27$ GFLOPs/tok, whereas Matryoshka and
Vanilla both reach $0.91$ GFLOPs/tok at th e 500M exit. By contrast,
Matryoshka places a 500M sub-model at the cumulative exit point of the
smallest block, with parameter count, FLOPs, and KV cache all scaling
down from there.

\paragraph{Matryoshka and KV Cache} On the FLOPs axis the three methods
collapse onto essentially the same curve as a function of parameters,
since FLOPs/tok is dominated by the same matrix-multiplication budget
regardless of how the parameters are distributed across attention and
FFN. KV cache, however, separates the methods cleanly.
\Cref{fig:matformer-kv} shows that MatFormer's KV cache is a horizontal
line at $280$ KB/tok regardless of which sub-model is extracted, because
attention shape and depth are fixed at the universal-model values. A
deployment that needs a smaller KV footprint, e.g.\ to fit longer
contexts or larger batches into the same memory budget, gains nothing
from MatFormer's elasticity. Matryoshka instead reduces KV cache from
$280$ KB to $96$ KB at the 500M exit, almost coincident with the
$96$ KB of an independently trained Vanilla 500M model. Matryoshka thus
matches the Vanilla curve on both axes across the full size range,
confirming that the nested structure does not introduce inference-time
overhead relative to a conventional suite, while delivering all
sub-models from a single training run.

\section{Out-of-Distribution Perplexity}
\label{sec:appendix-ood-ppl}

To probe how well each suite generalizes outside FineWeb-Edu, we evaluate
byte perplexity on five held-out corpora that span different domains:
WikiText-103 (encyclopedic), C4 (general web), PG-19 (long-form fiction),
arXiv (scientific full-text), and PubMed Central (biomedical full-text). All
runs use a 2\,048-token rolling window with stride 1\,024 and the same
SmolLM2 tokenizer. The Vanilla (FLOPs-matched) and Vanilla (Token-matched)
baselines mirror those in \Cref{tab:benchmarks}. We report byte perplexity
because it is invariant to tokenization differences and robust to long-tail
vocabulary in PG-19, arXiv and PubMed.

\Cref{tab:ood-ppl} reports the per-corpus byte perplexity. The advantage is
largest on arXiv, where the 3B Matryoshka sub-model lowers byte PPL by
$0.13$ over the Token-matched Vanilla baseline; the 1.5B and 500M Matryoshka
sub-models inherit part of this advantage through the shared early layers.

\begin{table}[h]
\centering\small
\begin{tabular}{lrrrrrr}
\toprule
Model & WikiText & C4 & PG-19 & arXiv & PubMed & \textbf{Average} \\
\midrule
\multicolumn{7}{l}{\textit{500M parameters}} \\
\midrule
Vanilla (FLOPs-matched) & 2.244          & 2.023          & 2.582          & 2.551          & 2.075          & 2.295 \\
Vanilla (Token-matched) & \textbf{2.215} & \textbf{2.003} & \textbf{2.504} & 2.530          & \textbf{2.065} & \textbf{2.263} \\
Matryoshka              & 2.221          & 2.010          & 2.509          & \textbf{2.497} & 2.090          & 2.265 \\
\midrule
\multicolumn{7}{l}{\textit{1.5B parameters}} \\
\midrule
Vanilla (FLOPs-matched) & 2.138          & 1.953          & 2.361          & 2.454          & 1.951          & 2.171 \\
Vanilla (Token-matched) & 2.107          & 1.934          & 2.283          & 2.447          & \textbf{1.925} & 2.139 \\
Matryoshka              & \textbf{2.100} & \textbf{1.931} & \textbf{2.228} & \textbf{2.398} & 1.946          & \textbf{2.121} \\
\midrule
\multicolumn{7}{l}{\textit{3B parameters}} \\
\midrule
Vanilla (FLOPs-matched) & 2.097          & 1.924          & 2.206          & 2.423          & 2.066          & 2.143 \\
Vanilla (Token-matched) & \textbf{2.059} & \textbf{1.906} & 2.129          & 2.442          & 1.950          & 2.097 \\
Matryoshka              & 2.064          & \textbf{1.906} & \textbf{2.120} & \textbf{2.315} & \textbf{1.931} & \textbf{2.067} \\
\bottomrule
\end{tabular}
\caption{Byte perplexity on five out-of-distribution corpora. Vanilla
(FLOPs-matched) and Vanilla (Token-matched) are defined as in
\Cref{tab:benchmarks}. Lower is better; bold marks the best within each
size group. Sample sizes are 1\,000 (WikiText-103), 200 (C4 and PubMed
Central), 40 (arXiv), and 20 (PG-19), all evaluated with a 2\,048-token
rolling window.}
\label{tab:ood-ppl}
\end{table}

\section{Closing the Per-Size Gap with Loss Weighting}
\label{sec:appendix-loss-weighting}

The default Matryoshka recipe weights each sub-model's loss equally in the total loss. This is a reasonable default but not
necessarily optimal: the 50M, 100M, and 200M sub-models do not contribute
equally to the gradient, and a uniform weighting over-serves some sizes at
the expense of others. We probe whether the residual gap to Vanilla
perplexity (\Cref{fig:dynamics:ppl}) can be closed by reweighting the
per-sub-model losses, using the 200M proxy suite.

We sweep nine weight vectors $(w_{50}, w_{100}, w_{200})$ chosen to redistribute
budget from sizes that are already at or below Vanilla toward those that
remain above it; all vectors are normalized to sum to one. We re-run the
last 1000 cooldown steps of the 200M Matryoshka run with each weighting,
keeping the rest of the recipe identical. The reference is the uniform
weighting $(1/3, 1/3, 1/3)$, which corresponds to the main recipe.

\begin{figure}[h]
    \centering
    \includegraphics[width=0.55\linewidth]{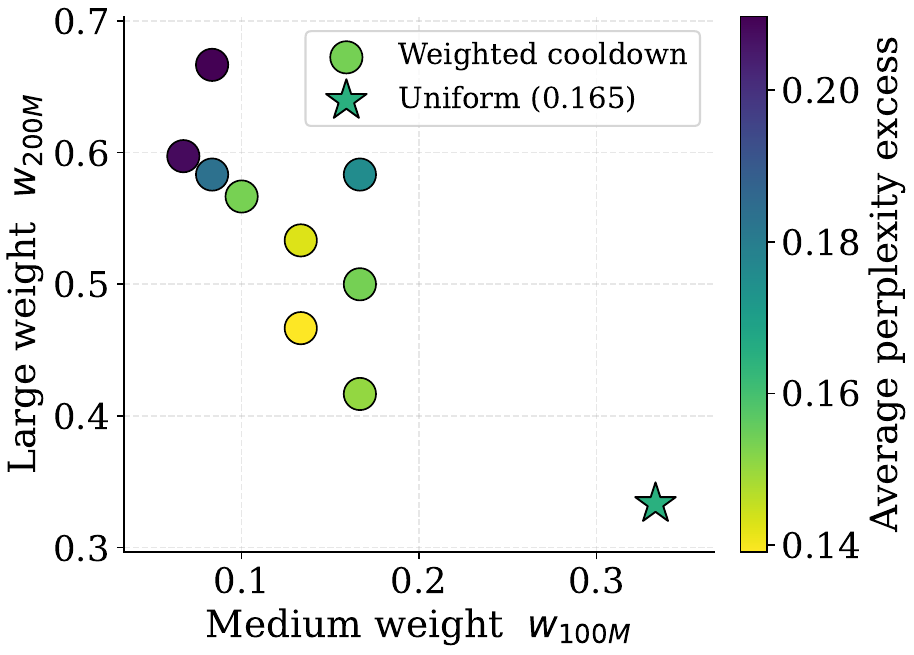}
    \caption{Average PPL excess over Vanilla (lower is better) on the 200M
      proxy suite for nine loss-weight configurations. Axes are the
      normalized weights of the medium (100M) and large (200M) sub-models;
      $w_{50}$ is the residual. The star marks the uniform weighting
      $(1/3, 1/3, 1/3)$, which is the main-recipe choice. Several weighted
      variants reduce the residual excess relative to uniform.}
    \label{fig:lw_excess}
\end{figure}

\Cref{fig:lw_excess} reports the average positive PPL excess of the
weighted runs over Vanilla, defined as $\mathbb{E}_{\text{size}}\left( \,
\max(0,\,\mathrm{PPL}_{\text{Mat.}}(\text{size}) -
\mathrm{PPL}_{\text{Van.}}(\text{size})) \right)$. This metric only penalizes sub-models that
remain above Vanilla, so it
captures \emph{how much negative gap remains} rather than rewarding improvements for
already-strong sub-models. Several weighted variants improve over uniform
on this aggregate score, with the best configuration cutting the residual
excess by roughly $15\%$.

\begin{figure}[h]
    \centering
    \begin{subcaptionblock}{0.27\linewidth}
        \includegraphics[width=\linewidth]{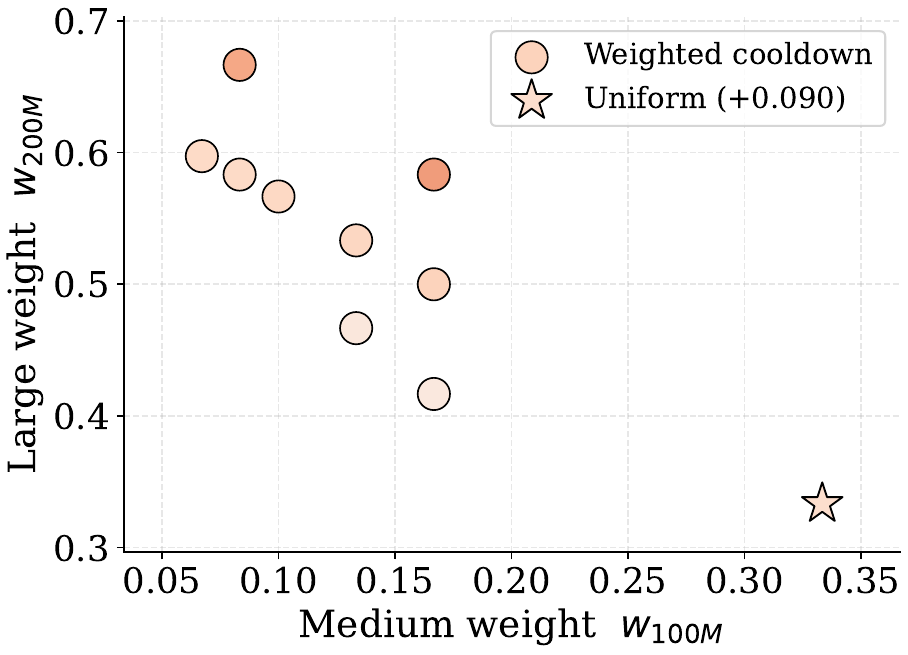}
        \caption{50M}
    \end{subcaptionblock}
    \hfill
    \begin{subcaptionblock}{0.27\linewidth}
        \includegraphics[width=\linewidth]{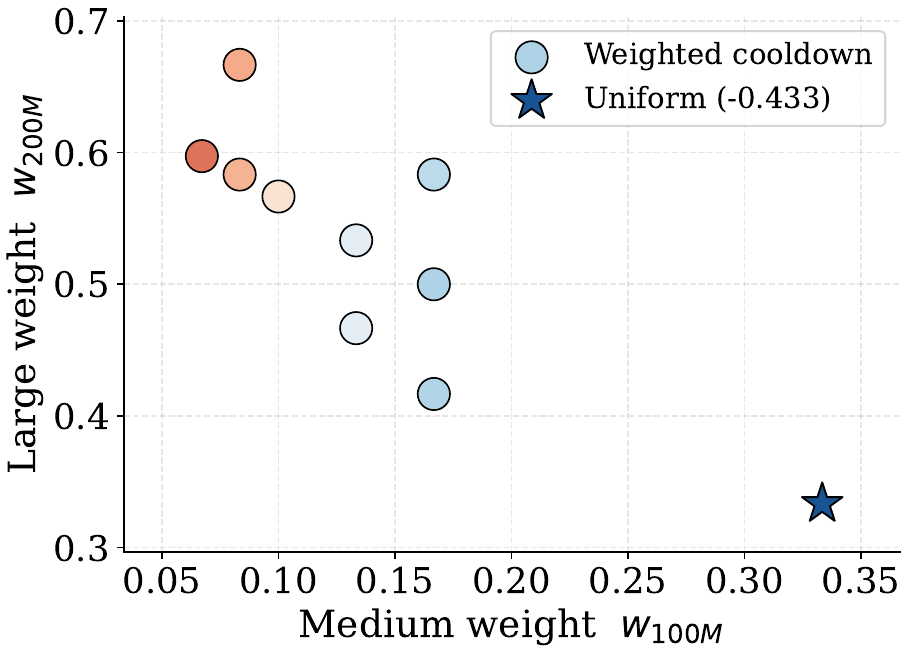}
        \caption{100M}
    \end{subcaptionblock}
    \hfill
    \begin{subcaptionblock}{0.27\linewidth}
        \includegraphics[width=\linewidth]{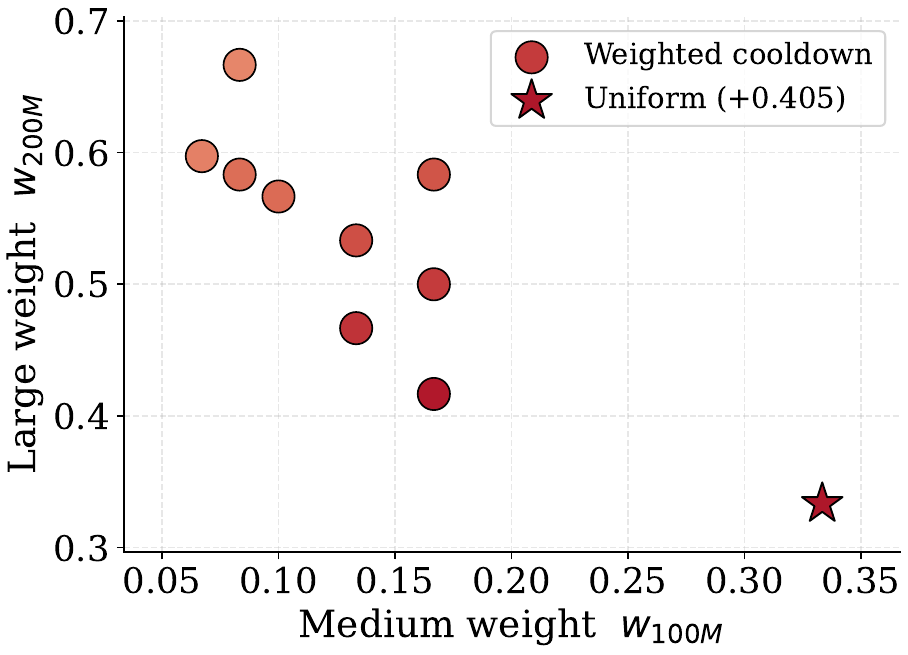}
        \caption{200M}
    \end{subcaptionblock}
    \hfill
    \begin{subcaptionblock}{0.06\linewidth}
        \includegraphics[width=\linewidth]{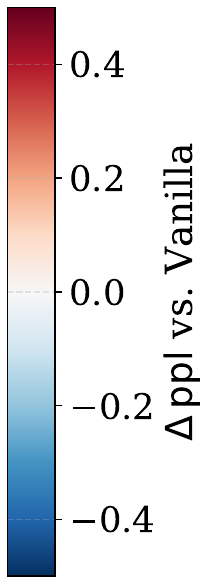}
        \vspace{1.5em}
    \end{subcaptionblock}
    \caption{Per-size $\Delta\mathrm{PPL}$ over Vanilla on the 200M proxy
      suite, as a function of the normalized weight vector $(w_{100},
      w_{200})$. Stars mark the uniform weighting. Blue points beat
      Vanilla; red points lag it. Under uniform weights the 100M sub-model
      is already below Vanilla while the 200M sub-model lags; redistributing
      weight from $w_{100}$ toward $w_{200}$ closes the 200M gap without
      breaking 100M.}
    \label{fig:lw_per_size}
\end{figure}

\Cref{fig:lw_per_size} disaggregates the same runs by sub-model size. Each
panel plots the per-size $\Delta\mathrm{PPL}$ over Vanilla as a function of
$(w_{100}, w_{200})$ (with $w_{50} = 1 - w_{100} - w_{200}$ implicit).
Two patterns are visible. First, under the uniform weighting (the star)
the 100M sub-model is already substantially below Vanilla while the 200M
sub-model is the bottleneck — equal weights over-serve 100M. Second,
weightings that redistribute budget from $w_{100}$ toward $w_{200}$ (and
mildly toward $w_{50}$) close the 200M gap and lift the 50M sub-model
above Vanilla without breaking the 100M sub-model. No single weighting is
Pareto-optimal on every axis, but several variants strictly dominate
uniform on the aggregate score and on the bottleneck sub-model.

We did not extend this sweep to the 3B suite, because its compute cost is
prohibitive and the qualitative finding is already useful at 200M. The
takeaway is that part of the residual per-size gap to Vanilla in the main
results is attributable to the uniform-weighting choice rather than to the
nested architecture. This suggests that a simple, near-zero-cost
hyperparameter tuning of $(w_m)$ would tighten the per-size results
further; we leave a systematic per-size loss-weighting rule, including
adaptive schedules, to future work.

\section{Additional Size-and-Shape Ablation Views}
\label{sec:appendix-size-shape}

\begin{figure}[h]
    \centering
    \begin{subcaptionblock}{0.32\linewidth}
        \includegraphics[width=\linewidth]{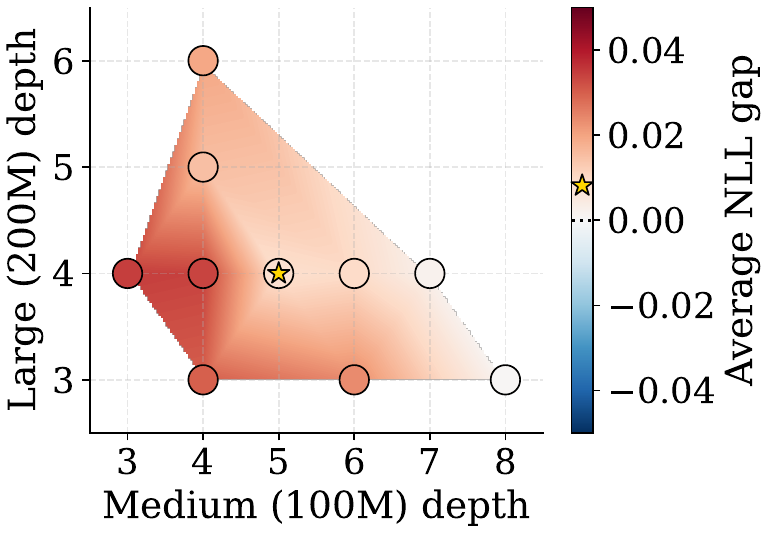}
        \caption{Mean $\Delta\text{NLL}$ scatter}
    \end{subcaptionblock}
    \hfill
    \begin{subcaptionblock}{0.32\linewidth}
        \includegraphics[width=\linewidth]{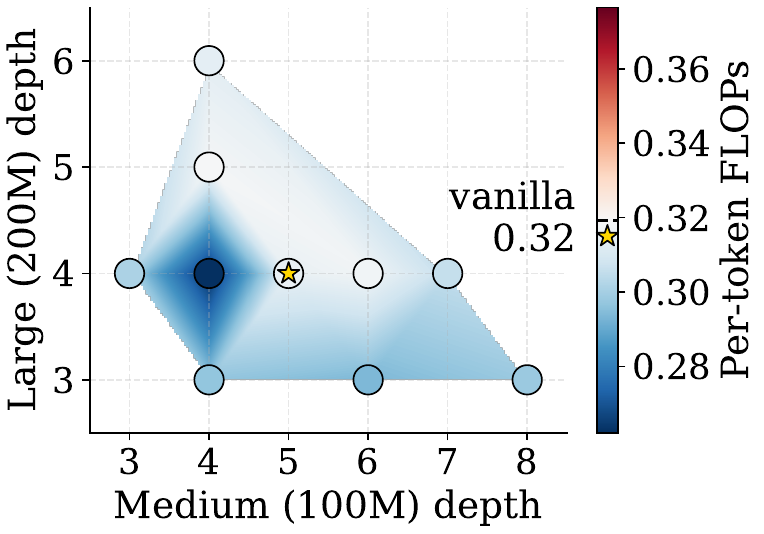}
        \caption{Per-token FLOPs (G)}
    \end{subcaptionblock}
    \hfill
    \begin{subcaptionblock}{0.32\linewidth}
        \includegraphics[width=\linewidth]{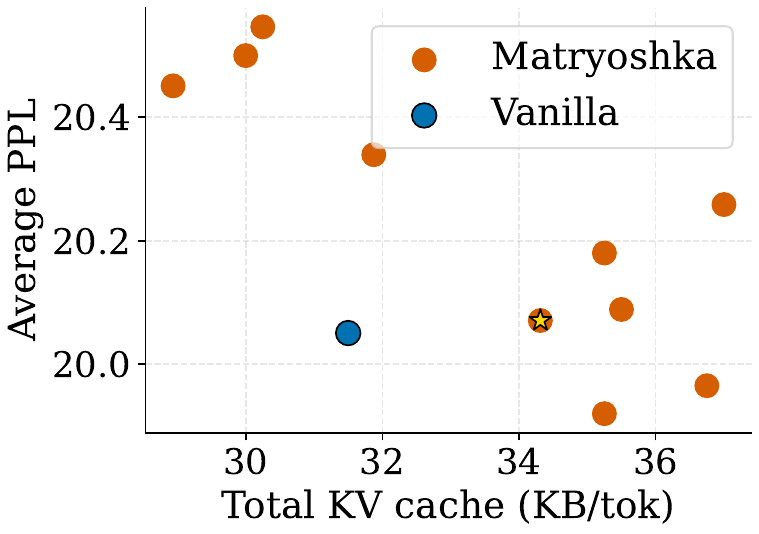}
        \caption{Mean NLL gap vs.\ KV cache}
    \end{subcaptionblock}
    \caption{Additional views of the 200M size-and-shape ablation.
      \textbf{(a)} Same axes as \Cref{fig:size_ablation}, colored by mean
      $\Delta\text{NLL}$ instead of $\Delta\text{ppl}$.
      \textbf{(b)} Per-token theoretical FLOPs of the full Matryoshka
      forward pass, with the Vanilla 200M baseline marked on the colorbar.
      \textbf{(c)} 1D summary plotting mean NLL gap (averaged over sub-model
      sizes) against total KV cache.}
    \label{fig:size_ablation_appendix}
\end{figure}

\Cref{fig:size_ablation} reports two views of the 200M size-and-shape sweep:
total KV cache per token and mean $\Delta\text{ppl}$ relative to a Vanilla
baseline. In \Cref{fig:size_ablation_appendix}, we report three additional
views: the mean $\Delta\text{NLL}$ counterpart of the perplexity scatter
(\Cref{fig:size_ablation_appendix}(a)), the per-token theoretical-FLOPs
scatter (\Cref{fig:size_ablation_appendix}(b)), and a 1D summary of mean NLL
gap as a function of total KV cache (\Cref{fig:size_ablation_appendix}(c)).
These views support the same conclusion as the main figure: the Matryoshka
design space contains variants that simultaneously reduce KV cache and FLOPs
relative to the Vanilla baseline without harming downstream quality.

As in \Cref{fig:size_ablation}, each point is one Matryoshka variant with a
shared 50M base sub-model, and $\Delta$ values are averaged over the three
sub-model sizes (50M / 100M / 200M) versus the Vanilla 200M baseline of
matching architecture.

\section{200M Proxy Suite Benchmarks}
\label{sec:appendix-200m-bench}

To complement the 3B-scale results, we evaluate the 200M proxy suite used
throughout our ablations on the same downstream benchmarks as
\Cref{tab:benchmarks}. The Vanilla baseline is the three independently
trained 50M, 100M, and 200M models that share the recipe and tokenizer of
the Matryoshka run; ``Matryoshka'' refers to our chosen v9 configuration.
Both suites are trained for 20B tokens.

In \Cref{tab:bench-200m}, the Matryoshka suite ties Vanilla on the 50M
sub-model ($-0.1$ avg) and beats it at 100M and 200M ($+0.4$ and $+0.1$
avg), confirming the perplexity parity reported in \Cref{fig:size_ablation_appendix} on downstream tasks.

\begin{table}[h]
\centering\small
\begin{tabular}{lccccccccc}
\toprule
Model & \multicolumn{1}{c}{ARC-E} & \multicolumn{1}{c}{ARC-C} & \multicolumn{1}{c}{HS} & \multicolumn{1}{c}{Lambada} & \multicolumn{1}{c}{OBQA} & \multicolumn{1}{c}{PIQA} & \multicolumn{1}{c}{WG} & \multicolumn{1}{c}{\textbf{Avg acc.} $\uparrow$} \\
\midrule
\multicolumn{9}{l}{\textit{50M parameters}} \\
\midrule
Vanilla    & \textbf{44.2} & 23.6          & \textbf{28.5} & \textbf{17.5} & \textbf{29.4} & \textbf{60.1} & 50.2          & \textbf{36.2} \\
Matryoshka & 43.1          & \textbf{24.6} & 27.9          & 16.8          & 29.0          & 59.6          & \textbf{51.7} & 36.1          \\
\midrule
\multicolumn{9}{l}{\textit{100M parameters}} \\
\midrule
Vanilla    & \textbf{47.4} & 25.4          & \textbf{32.3} & 24.1          & 31.0          & 62.1          & 51.5          & 39.1          \\
Matryoshka & 47.2          & \textbf{25.5} & 32.1          & \textbf{24.9} & \textbf{32.0} & \textbf{62.7} & \textbf{52.0} & \textbf{39.5} \\
\midrule
\multicolumn{9}{l}{\textit{200M parameters}} \\
\midrule
Vanilla    & \textbf{51.8} & \textbf{27.7} & \textbf{36.7} & 29.4          & \textbf{33.2} & 64.5          & 50.8          & 42.0          \\
Matryoshka & 51.6          & 26.7          & 36.2          & \textbf{30.0} & 32.2          & \textbf{65.9} & \textbf{52.1} & \textbf{42.1} \\
\bottomrule
\end{tabular}
\caption{Per-benchmark zero-shot accuracy on the 200M proxy suite (50M /
100M / 200M sub-models). Vanilla is three independently trained models;
Matryoshka is the configuration used as the reference point throughout
our ablations. Both suites are trained for 20B tokens. Avg acc.\ averages
the seven benchmarks (WG = Winogrande). Bold marks the better entry per
row group.}
\label{tab:bench-200m}
\end{table}


\end{document}